\documentclass{article}

\usepackage{arxiv}

\usepackage[utf8]{inputenc} 
\usepackage[T1]{fontenc}    
\usepackage{hyperref}       
\hypersetup{
	colorlinks   = true, 
	urlcolor     = blue, 
	linkcolor    = blue, 
	citecolor   = blue 
}
\usepackage{url}            
\usepackage{booktabs}       
\usepackage{amsfonts}       
\usepackage{nicefrac}       
\usepackage{microtype}      
\usepackage{lipsum}
\usepackage{graphicx}
\usepackage{amsmath}
\usepackage{amsfonts}
\usepackage{amssymb}
\usepackage{multirow}
\usepackage{float}
\usepackage{adjustbox}
\usepackage{booktabs}
\usepackage{lscape}
\usepackage{enumitem, array}
\usepackage{xcolor}
\usepackage[round,authoryear]{natbib}
\usepackage{algorithm2e}
\usepackage{subcaption}
\SetAlgorithmName{Procedure}{procedure}{List of Procedures}

\usepackage{tikz}
\usepackage{bm}
\usepackage{relsize}

\usetikzlibrary{positioning}

\title{Data-Driven System Identification of \\6-DoF Ship Motion in Waves with Neural Networks}

\author{
	Kevin M. Silva$^{1,2,\star}$ and Kevin J. Maki$^{2}$ \\
	\\
	$^1$Naval Surface Warfare Center Carderock Division, USA\\
	$^2$Department of Naval Architecture and Marine Engineering, The University of Michigan, USA\\
	$^\star$Corresponding author: \texttt{kevin.m.silva1@navy.mil} \\
}

\begin{document}
	\maketitle
	
	\begin{abstract}
		Critical evaluation and understanding of ship responses in the ocean is important for not only the design and engineering of future platforms but also the operation and safety of those that are currently deployed. Simulations or experiments are typically performed in nominal sea conditions during ship design or prior to deployment, to evaluate different hullforms and develop operational profiles. Though it is necessary to analyze the response for nominal conditions, the results may not be reflective of the instantaneous state of the vessel and the ocean environment while deployed. Short-term temporal predictions of ship responses given the current wave environment and ship state would enable enhanced decision-making onboard and reduce the overall risk during operations for both manned and unmanned vessels, especially as the marine industry trends towards more autonomy. However, the current state-of-the-art in numerical hydrodynamic simulation tools are too computationally expensive to be employed for real-time ship motion forecasting and the computationally efficient tools are too low fidelity to provide accurate responses. Thus, a methodology is needed to provide fast and efficient predictions with levels of accuracy closer to the higher-fidelity tools.A methodology is developed with long short-term memory (LSTM) neural networks to represent the motions of a free running David Taylor Model Basin (DTMB) 5415 destroyer operating at 20 knots in Sea State 7 stern-quartering irregular seas. Case studies are performed for both course-keeping and turning circle scenarios. An estimate of the vessel's encounter frame is made with the trajectories observed in the training dataset. Wave elevation time histories are given by artificial wave probes that travel with the estimated encounter frame and serve as input into the neural network, while the output is the 6-DOF temporal ship motion response. Overall, the neural network is able to  predict the temporal response of the ship due to unseen waves accurately, which makes this methodology suitable for system identification and real-time ship motion forecasting. The methodology, the dependence of model accuracy on wave probe and training data quantity and the estimated encounter frame are all detailed.
	\end{abstract}

	
	\section*{Introduction}
		
	The ability to predict ship responses in waves accurately and efficiently is a challenging problem. Ships operate at various speeds and headings and in harsh ocean environments where different wave excitation can lead to severe responses that can harm not only the ship but personnel onboard. Therefore, proper quantification and understanding of the responses due to different wave environments is paramount. Evaluations and predictions are performed by engineers and designers with a variety of numerical hydrodynamic tools. These evaluations can rely on numerical predictions from frequency-domain strip-theory formulations like those developed by \cite{Salvesen1970}, blended nonlinear time-domain methods such as the Large Amplitude Motion Program (LAMP) \citep{Lin2007} that leverages a panel method, and TEMPEST \citep{Belknap2019} which utilizes strip-theory to solve the nonlinear hydrostatics and hydrodynamics, as well as sophisticated force models to model green water and viscous effects. More recently, high-fidelity computational fluid dynamic (CFD) tools with Unsteady Reynolds-averaged Navier-Stokes (URANS) formulations have become more prevalent when predicting ship motions in waves. URANS simulations have been included in the development of force models for potential flow tools like the work of \cite{Aram2019}, capsizing and broaching in regular waves \citep{Hosseini2009,Mousaviraad2010}, or within probabilistic frameworks to observe extreme events and calculate the probability of their occurence like the work of \cite{Xu2020} for the Design Loads Generator (DLG) method or \cite{Silva2021oe} for the critical wave groups (CWG) method.
	
	A large focus recently has been placed on the prediction of the six degree-of-freedom (DoF) response of vessels in waves where the horizontal plane motions such as surge, sway and yaw can vary greatly. This contrasts with classical seakeeping evaluations, where the vessel is assumed to travel at a constant speed and heading. Previous studies have shown success in predicting 6-DoF motions with URANS simulations {\citep{Serani2021}}, while others have employed a potential flow time-domain solution \citep{Lin2006}, or a hybrid formulation in \cite{White2021} where a URANS double-body formulation and potential flow are combined in a tightly-coupled solver to reduce the computational cost of a volume-of-fluid (VOF) URANS simulation, yet still model some of the important viscous features.
	
	Although predictions of 6-DoF ship responses can be made with URANS or potential flow simulations, their computational cost can be prohibitive when long exposure windows are needed or faster than real-time predictions are required for applications such as ship motion forecasting, maneuvering simulators for crew training, and development of control systems. Thus, analyses with these numerical hydrodynamic tools must be performed for many realizations of wave sequences in a nominal seaway to develop a statistical description of the response. A statistical description of the ship response is crucial to design better and safer ships and also to guide the operation of existing platforms away from dangerous situations and towards more favorable ones. However, due to the computational cost of the hydrodynamic tools, evaluations are conducted in nominal seaways and may not be representative of the instantaneous state of the vessel and ocean environment. Though these evaluations provide response statistics based on nominal operating and environmental conditions, they lack the ability to provide the crew upon manned vessels and autonomous systems onboard unmanned vessels with real-time forecasting of ship responses, given the current environment. Accurate forecasting of the vessel's temporal response would allow for both manned and unmanned vessels to operate safer and push the boundaries of their operational envelope with higher confidence as the different sectors of the marine industry push towards more autonomy.
	
	The current paper explores the system identification (SI) of a free running ship in waves. SI is used to develop a surrogate model of a complex process. In the context of ship responses in waves, the goal of the SI is to create a fast-running model capable of representing either simulations with a higher-fidelity numerical hydrodynamic tool, model test, or full-scale data. The resulting model can run conditions not considered in its development to expand the analysis to new conditions. In prediction of ship motions in waves, different implementations of SI have been considered, ranging from coefficient-based mathematical models developed in \cite{Araki2019} where parameters were tuned with CFD simulations to neural network-based models. \cite{Hess2007} presented a method with Recurrent Neural Networks (RNN) that attempted to predict the 6-DoF motions of a full-scale ship operating in random waves. Though \cite{Hess2007} was able to produce predictions that were qualitatively similar for some validation cases, larger discrepancies occurred in other cases and overall the RNN model did not produce results that were quantitatively representative of the underlying dynamics. The work of \cite{Xu2021} showcased the ability of long short-term memory (LSTM) neural networks to predict the nonlinear propagation of a wave field downstream of a wave probe as well as the heave and roll of a midship section predicted by CFD. Also, \cite{Ferrandis2021} applied LSTM and gated recurrent unit (GRU) neural networks to predict motion of vessel traveling at a constant speed and heading. Additionally, LSTM neural networks have been utilized by \cite{Silva2021marine, Silva2021stab} to act as a surrogate for CFD predictions of the extreme roll of a midship section within an the critical wave groups (CWG) extreme event probabilistic method. The work of \cite{Xu2021, Silva2021marine, Silva2021stab, Ferrandis2021} all showcase the ability of either LSTM or GRU neural networks to represent ship motions due to wave excitation but are limited in that they restrict the DoF in the horizontal plane (surge, sway, and yaw). The case studies in \cite{Xu2021, Silva2021marine, Silva2021stab} were performed at zero speed and constrained sway and yaw, while the case studies in \cite{Ferrandis2021} considered constant speed and heading. Though the assumption of constant speed and heading may be sufficient for some cases, significant variation in the horizontal plane responses can affect the other DoF and overall global response of the vessel.
	
	The surrogate models developed through SI techniques can be implemented within a forecasting framework where real-time vessel and wave information can be leveraged to provide faster than real-time predictions of future ship responses. For accurate forecasting predictions, a complete 6-DoF representation  of the ship response is required. Recent work by \cite{DAgostino2021} compared the ability of different neural network architectures to produce a nowcasting encoder-decoder model capable of predicting a short-term window of the response. Though the models were successful in some DoF, surge and sway produced undesirable results and predictions of all DoF became worse as time increased. \cite{Diez2021} proposed a dynamic mode decomposition method for the forecasting of ship motions but similar to \cite{DAgostino2021}, the predictions were close initially but largely drifted as time progressed. The previous work with neural networks has showcased their efficacy when representing simplified ship motions but expansion to a fully free-running ship with implications for nowcasting and forecasting has not been completely developed.
	
	The objective of the current work is to extend the work performed in \cite{Xu2021} and develop a modeling approach with an LSTM neural network to represent the 6-DoF response of a vessel in waves accurately and also quantify the uncertainty of the predicted temporal response. The resulting model will not only be suitable for SI but also works within a ship motion forecasting framework. The remainder of the paper is organized as follows: The proposed modeling approach is illustrated with details about the data-preparation, neural network architecture, training, inference, and uncertainty quantification. Finally, the model is evaluated for the representation of the 6-DoF response predicted by LAMP of the David Taylor Model Basin (DTMB) 5415 hullform \citep{Longo2005} in stern-quartering Sea State 7 \cite{NATO1983} irregular seas and self-propelled at 20 knots for both course-keeping and turning circle cases, with a proportional–integral–derivative (PID) controller prescribing rudder motion to maintain heading during course-keeping.  The developed models are evaluated for their accuracy and convergence with respect to number of wave probes and training data quantity.

    \section*{Methodology}
	
	Different approaches have been taken to develop data-driven models that represent ship motions. One approach is to develop models that predict statistics of the response such as the standard deviation \citep{Schirmann2020} or produce operational guidance envelopes that are based off ensemble statistics developed through simulations, model tests, or full-scale measurements. However, these approaches do not provide insight into the mechanisms of a particular response and are not entirely useful for motion forecasting in the short-term as longer exposure windows are needed for statistics to be satisfied. Therefore, development of an SI model that can provide an accurate temporal response is highly desirable. The general objective of an SI methodology is to develop a model that given an input, can produce an output that is representative of the underlying system of interest. In the case of a causal dynamical system (e.g. ship motions in waves), the output is not only dependent on the current excitation but also depends on previous excitations as well. At any given time index $t$, the output of a discrete dynamical system $y_t$ can be characterized by the input at the current time index $x_t$ and the input at previous time indices $\left( x_{t-1}, x_{t-2},...\right)$. Eqn.~(\ref{eq:reg_state}) shows the calculation of a state variable $s_t$, at the current time index, using the inputs and state variables at previous time indices and the mapping function $f$. Eqn.~(\ref{eq:reg_output}) shows the mapping of the state variables to the output of the dynamical system with the mapping function $g$. For simple dynamical systems, the mapping $f$ may be clear, but for more complicated systems, machine learning models are employed to solve a regression problem and identify the best nonlinear mappings between both the input time series $\left( x_{t-1}, x_{t-2},...\right)$ and state variables $\left( s_{t-1}, s_{t-2},...\right)$, and the state variables and output time series $\left( y_{t-1}, y_{t-2},...\right)$.
		
	\begin{equation}
	s_t = f(s_{t-1}, s_{t-2}, \cdots; x_t, x_{t-1}, \cdots)
	\label{eq:reg_state}
	\end{equation}
		
	\begin{equation}
	y_t = g(s_t, s_{t-1}, s_{t-2}, \cdots)
	\label{eq:reg_output}
	\end{equation}

	\subsection*{Neural Networks}
	    
\begin{figure}[H]
		\centering
		\includegraphics[width=0.37\textwidth]{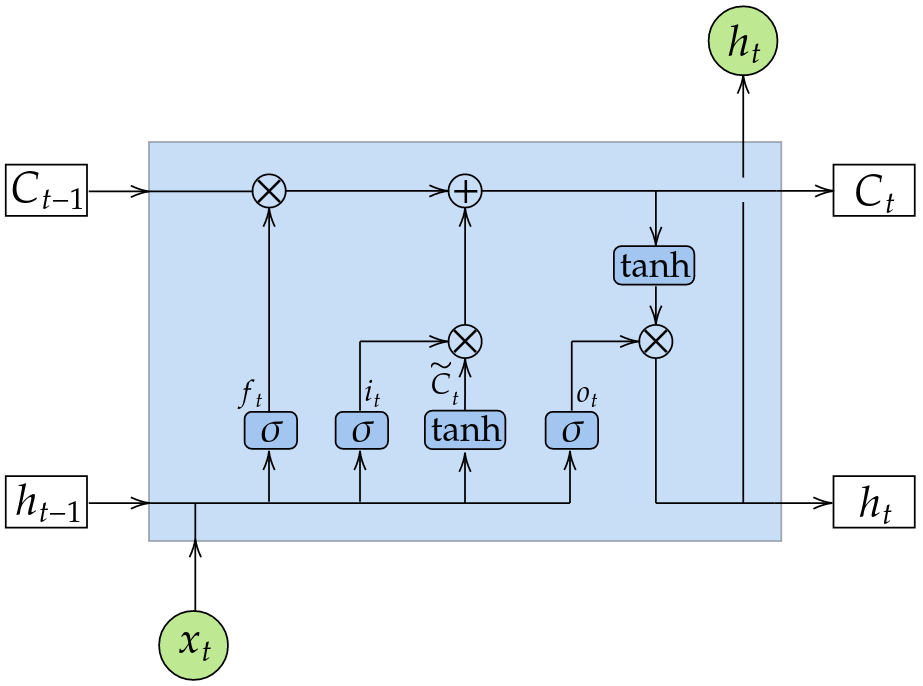}
		\caption{Layout of an LSTM unit.}
		\label{fig:lstm}
	\end{figure}

	The present work leverages LSTM neural networks \citep{Hochreiter1997, Olah2015}, an implementation of the RNN family of neural networks \citep{Hopfield1982, Rumelhart1986}. LSTM neural networks have shown historical success in sequence learning type tasks such as natural language processing and speech recognition. LSTM neural networks have also been popular in solving marine dynamics problems such as nonlinear wave propagation and ship motions due to wave excitation. Fig.~\ref{fig:lstm} shows the layout of an LSTM unit and demonstrates how given a cell input $x_t$, the previous output $h_{t-1}$, and the previous cell state $C_{t-1}$, the current cell state $C_t$ and current cell output $h_t$ can be calculated. Eqn.~(\ref{eq:f_t}) through (\ref{eq:h_t}) demonstrate how all the various parameters within the LSTM cell are calculated where, $\sigma(\cdot)$ corresponds to the sigmoid function, and $W_f$, $b_f$, $W_i$, $b_i$, $W_C$, $b_C$, $W_o$, and $b_o$ are the tunable parameters of the LSTM that are learned throughout the training process.

	\begin{equation}
		f_t = \sigma(W_f\underbrace{[h_{t-1}, x_t]}_{\text{concatenate}} + b_f)
		\label{eq:f_t}
	\end{equation}
	
	\begin{equation}
		i_t = \sigma(W_i\underbrace{[h_{t-1}, x_t]}_{\text{concatenate}} + b_i)
		\label{eq:i_t}
	\end{equation}
	
	\begin{equation}
		\tilde{C}_t = \tanh(W_C\underbrace{[h_{t-1}, x_t]}_{\text{concatenate}} + b_C)
		\label{eq:tilde_C_t}
	\end{equation}
	
	\begin{equation}
		C_t = f_t * C_{t-1} + i_t * \tilde{C}_t
		\label{eq:C_t}
	\end{equation}
	
	\begin{equation}
		o_t = \sigma(W_o\underbrace{[h_{t-1}, x_t]}_{\text{concatenate}} + b_o)
		\label{eq:o_t}
	\end{equation}

    \begin{equation}
	    h_t = o_t * \tanh(C_t)
		\label{eq:h_t}
	\end{equation}

\begin{figure}[H]
		\centering
		\includegraphics[width=0.4\textwidth]{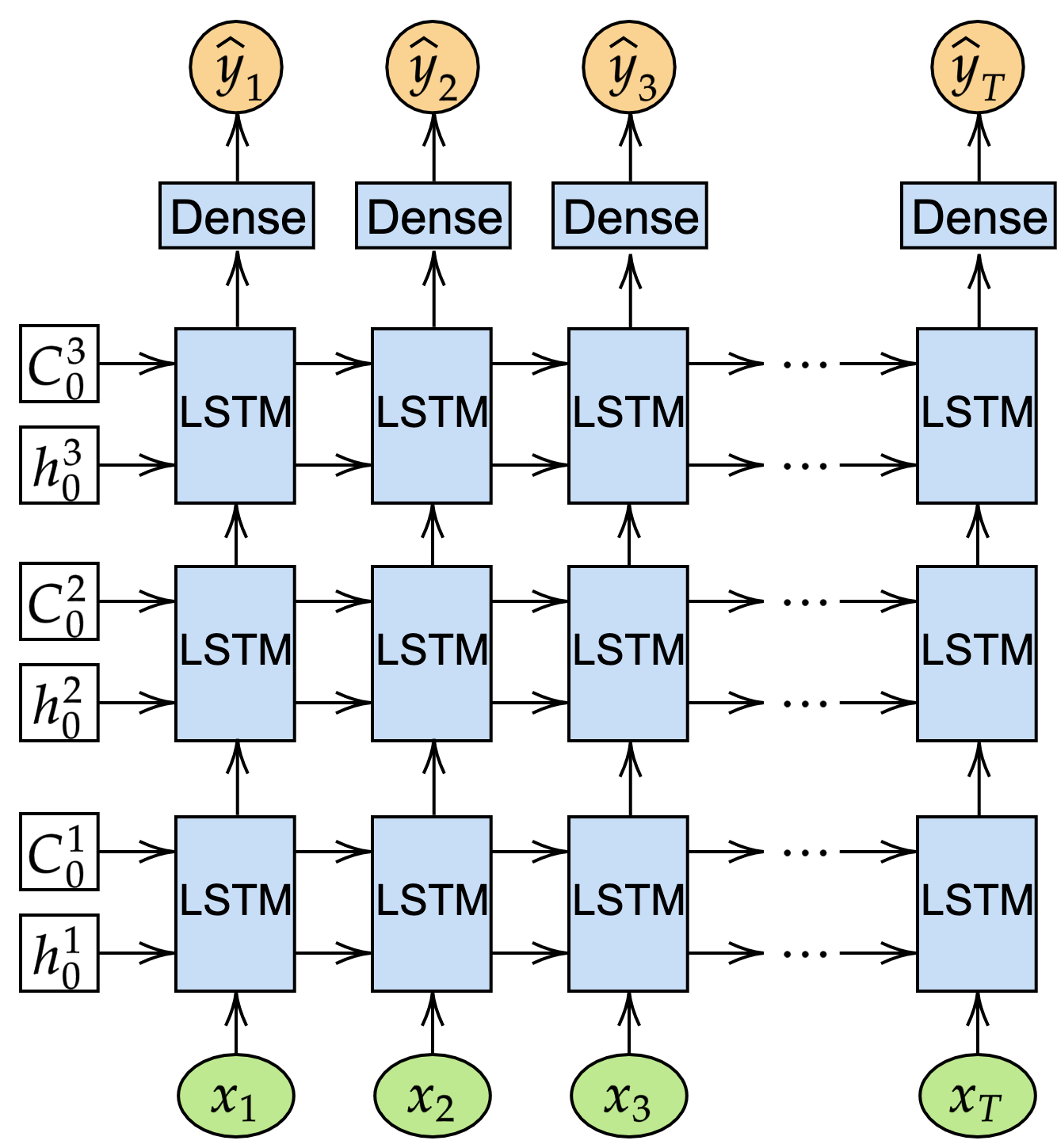}
		\caption{Neural network architecture.}
		\label{fig:nn-arch}
	\end{figure}

 Fig.~\ref{fig:nn-arch} demonstrates the neural network architecture implemented in the present paper in an unfolded view. The proposed architecture is similar to those in \cite{Xu2021,Silva2021marine,Silva2021stab}, except that three LSTM layers are utilized in the current work. In Fig.~\ref{fig:nn-arch}, the input layer is followed by three LSTM layers where the output of each previous layer becomes the input to the subsequent layer. The last LSTM layer is then followed by a fully-connected dense layer with linear activation which generates the temporal output of the neural network.

	\subsection*{Uncertainty}
	
\begin{figure}[H]
\centering
\begin{tikzpicture}

	\node[circle, draw, thick] (i1) {};
	\node[circle, draw, thick, above=2em of i1] (i2) {};
	\node[circle, draw, thick, above=2em of i2] (i3) {};
	\node[circle, draw, thick, below=2em of i1] (i4) {};
	\node[circle, draw, thick, below=2em of i4] (i5) {};
	
	\node[circle, draw, thick, right=4em of i1] (h1) {};
	\node[circle, draw, thick, right=4em of i2] (h2) {};
	\node[circle, draw, thick, right=4em of i3] (h3) {};
	\node[circle, draw, thick, right=4em of i4] (h4) {};
	\node[circle, draw, thick, right=4em of i5] (h5) {};
	
	\node[circle, draw, thick, right=4em of h1] (hh1) {};
	\node[circle, draw, thick, right=4em of h2] (hh2) {};
	\node[circle, draw, thick, right=4em of h3] (hh3) {};
	\node[circle, draw, thick, right=4em of h4] (hh4) {};
	\node[circle, draw, thick, right=4em of h5] (hh5) {};
	
	\node[circle, draw, thick, right=4em of hh2] (o1) {};
	\node[circle, draw, thick, right=4em of hh4] (o2) {};
	
	\draw[-stealth, thick] (i1) -- (h1);
	\draw[-stealth, thick] (i1) -- (h2);
	\draw[-stealth, thick] (i1) -- (h3);
	\draw[-stealth, thick] (i1) -- (h4);
	\draw[-stealth, thick] (i1) -- (h5);
	\draw[-stealth, thick] (i2) -- (h1);
	\draw[-stealth, thick] (i2) -- (h2);
	\draw[-stealth, thick] (i2) -- (h3);
	\draw[-stealth, thick] (i2) -- (h4);
	\draw[-stealth, thick] (i2) -- (h5);
	\draw[-stealth, thick] (i3) -- (h1);
	\draw[-stealth, thick] (i3) -- (h2);
	\draw[-stealth, thick] (i3) -- (h3);
	\draw[-stealth, thick] (i3) -- (h4);
	\draw[-stealth, thick] (i3) -- (h5);
	\draw[-stealth, thick] (i4) -- (h1);
	\draw[-stealth, thick] (i4) -- (h2);
	\draw[-stealth, thick] (i4) -- (h3);
	\draw[-stealth, thick] (i4) -- (h4);
	\draw[-stealth, thick] (i4) -- (h5);
	\draw[-stealth, thick] (i5) -- (h1);
	\draw[-stealth, thick] (i5) -- (h2);
	\draw[-stealth, thick] (i5) -- (h3);
	\draw[-stealth, thick] (i5) -- (h4);
	\draw[-stealth, thick] (i5) -- (h5);
	
	\draw[-stealth, thick] (h1) -- (hh1);
	\draw[-stealth, thick] (h1) -- (hh2);
	\draw[-stealth, thick] (h1) -- (hh3);
	\draw[-stealth, thick] (h1) -- (hh4);
	\draw[-stealth, thick] (h1) -- (hh5);
	\draw[-stealth, thick] (h2) -- (hh1);
	\draw[-stealth, thick] (h2) -- (hh2);
	\draw[-stealth, thick] (h2) -- (hh3);
	\draw[-stealth, thick] (h2) -- (hh4);
	\draw[-stealth, thick] (h2) -- (hh5);
	\draw[-stealth, thick] (h3) -- (hh1);
	\draw[-stealth, thick] (h3) -- (hh2);
	\draw[-stealth, thick] (h3) -- (hh3);
	\draw[-stealth, thick] (h3) -- (hh4);
	\draw[-stealth, thick] (h3) -- (hh5);
	\draw[-stealth, thick] (h4) -- (hh1);
	\draw[-stealth, thick] (h4) -- (hh2);
	\draw[-stealth, thick] (h4) -- (hh3);
	\draw[-stealth, thick] (h4) -- (hh4);
	\draw[-stealth, thick] (h4) -- (hh5);
	\draw[-stealth, thick] (h5) -- (hh1);
	\draw[-stealth, thick] (h5) -- (hh2);
	\draw[-stealth, thick] (h5) -- (hh3);
	\draw[-stealth, thick] (h5) -- (hh4);
	\draw[-stealth, thick] (h5) -- (hh5);

	\draw[-stealth, thick] (hh1) -- (o1);
	\draw[-stealth, thick] (hh1) -- (o2);
	\draw[-stealth, thick] (hh2) -- (o1);
	\draw[-stealth, thick] (hh2) -- (o2);
	\draw[-stealth, thick] (hh3) -- (o1);
	\draw[-stealth, thick] (hh3) -- (o2);
	\draw[-stealth, thick] (hh4) -- (o1);
	\draw[-stealth, thick] (hh4) -- (o2);
	\draw[-stealth, thick] (hh5) -- (o1);
	\draw[-stealth, thick] (hh5) -- (o2);
	
	Before
	
	\draw[-stealth, thick] (5,0) -- node[above] {Apply Dropout} (7.2, 0);

	
	\node[circle, draw, thick, right=10em of hh1] (i1) {};
	\node[circle, draw, thick, above=2em of i1] (i2) {};
	\node[circle, draw, thick, above=2em of i2] (i3) {};
	\node[circle, draw, thick, below=2em of i1] (i4) {};
	\node[circle, draw, thick, below=2em of i4] (i5) {};
	
	\node[circle, draw, thick, red, fill=red!10, right=4em of i1] (h1) {};
	\node[circle, draw, thick, right=4em of i2] (h2) {};
	\node[circle, draw, thick, red, fill=red!10, right=4em of i3] (h3) {};
	\node[circle, draw, thick, red, fill=red!10, right=4em of i4] (h4) {};
	\node[circle, draw, thick, right=4em of i5] (h5) {};
	
	\node[red] (icr) at (h1) {$\mathlarger{\mathlarger{\mathlarger{\mathlarger{\mathlarger{\bm{\times}}}}}}$};
	\node[red] (icr) at (h3) {$\mathlarger{\mathlarger{\mathlarger{\mathlarger{\mathlarger{\bm{\times}}}}}}$};
	\node[red] (icr) at (h4) {$\mathlarger{\mathlarger{\mathlarger{\mathlarger{\mathlarger{\bm{\times}}}}}}$};
	
	\node[circle, draw, thick, right=4em of h1] (hh1) {};
	\node[circle, draw, thick, red, fill=red!10, right=4em of h2] (hh2) {};
	\node[circle, draw, thick, right=4em of h3] (hh3) {};
	\node[circle, draw, thick, red, fill=red!10, right=4em of h4] (hh4) {};
	\node[circle, draw, thick, right=4em of h5] (hh5) {};
	
	\node[red] (icr) at (hh2) {$\mathlarger{\mathlarger{\mathlarger{\mathlarger{\mathlarger{\bm{\times}}}}}}$};
	\node[red] (icr) at (hh4) {$\mathlarger{\mathlarger{\mathlarger{\mathlarger{\mathlarger{\bm{\times}}}}}}$};
	
	\node[circle, draw, thick, right=4em of hh2] (o1) {};
	\node[circle, draw, thick, right=4em of hh4] (o2) {};
	
	\draw[-stealth, thick] (i1) -- (h2);
	\draw[-stealth, thick] (i1) -- (h5);
	\draw[-stealth, thick] (i2) -- (h2);
    \draw[-stealth, thick] (i2) -- (h5);
	\draw[-stealth, thick] (i3) -- (h2);
	\draw[-stealth, thick] (i3) -- (h5);
	\draw[-stealth, thick] (i4) -- (h2);
	\draw[-stealth, thick] (i4) -- (h5);
	\draw[-stealth, thick] (i5) -- (h2);
	\draw[-stealth, thick] (i5) -- (h5);
	
	\draw[-stealth, thick] (h2) -- (hh1);
	\draw[-stealth, thick] (h2) -- (hh3);
	\draw[-stealth, thick] (h2) -- (hh5);
	\draw[-stealth, thick] (h5) -- (hh1);
	\draw[-stealth, thick] (h5) -- (hh3);
	\draw[-stealth, thick] (h5) -- (hh5);
	
	\draw[-stealth, thick] (hh1) -- (o1);
	\draw[-stealth, thick] (hh1) -- (o2);
	\draw[-stealth, thick] (hh3) -- (o1);
	\draw[-stealth, thick] (hh3) -- (o2);
	\draw[-stealth, thick] (hh5) -- (o1);
	\draw[-stealth, thick] (hh5) -- (o2);

\end{tikzpicture}
\caption{Application of dropout regularization technique.}
\label{fig:dropout}
\end{figure}
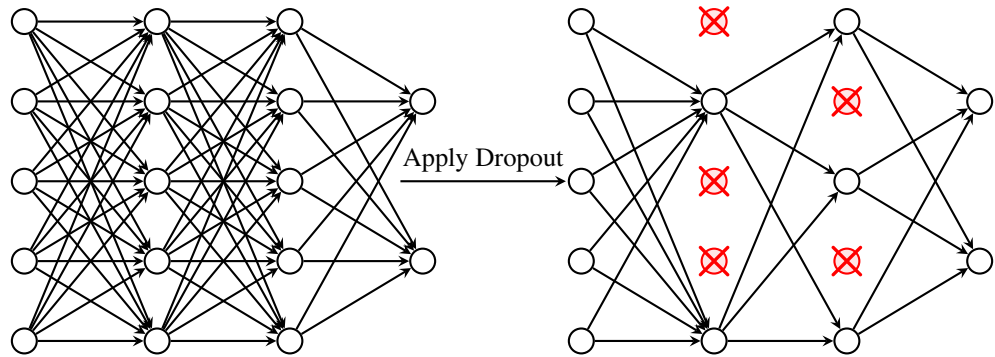

    Machine learning and neural networks in particular have gained a tremendous amount of popularity in recent years and are being implemented across several disciplines in both academia and industry. Though much focus has been placed on development of the most accurate models that describe a dataset, knowing when the models are not applicable can be just as or even more important. Thus, the quantification of uncertainty in neural network predictions is a necessity in any of their applications. The present work utilizes the Monte Carlo Dropout approach developed by \cite{Gal2016a,Gal2016b} to quantify uncertainty. Dropout is a common regularization technique employed in neural networks where a portion of the neurons are probabilistically excluded from activation and weight updates while training the model. This technique, demonstrated in Fig.~\ref{fig:dropout}, prevents the model from over-fitting as well as propels the optimization scheme towards a more generalized model that is capable of producing accurate predictions outside of the training dataset.

\begin{figure}[H]
	\centering
	\includegraphics[width=0.6\textwidth]{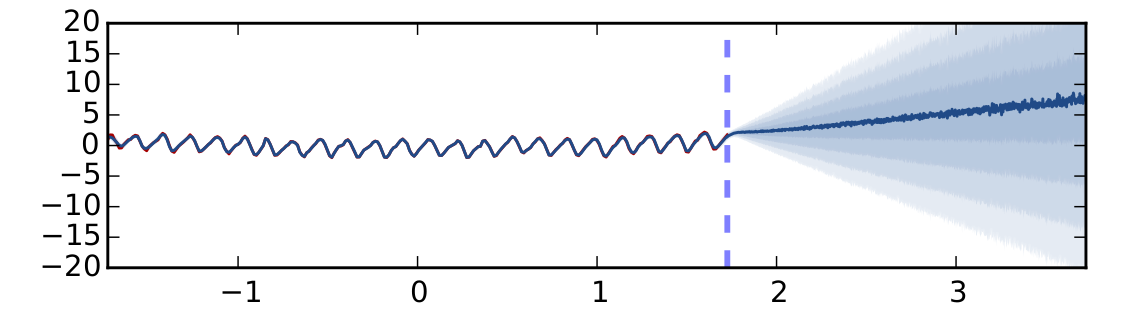}
	\caption{Demonstration of Monte Carlo dropout from \cite{Gal2016a}.}
	\label{fig:galunc}
\end{figure}

    \cite{Gal2016a,Gal2016b} proposed a Monte Carlo dropout approach, where dropout is also performed during prediction, thus providing an ensemble of predictions. This ensemble of predictions could then be leveraged to provide uncertainty estimates. In the context of the current work, a dropout layer is added after each LSTM layer to implement the Monte Carlo dropout methodology .This type of uncertainty estimate is extremely useful as it does not require any intrusive modification to the neural network architecture other than adding dropout. Previous work with the Monte Carlo dropout has demonstrated its effectiveness in providing large uncertainty estimates when a model is used to make a prediction that is outside of the training dataset. Fig.~\ref{fig:galunc} shows an example with the Monte Carlo dropout approach for uncertainty. The observed function in red is shown to the left of the dashed line, while the predictive mean plus/minus two standard deviations is shown in blue. With the Monte Carlo dropout approach, the model is able to denote an area of large uncertainty outside of what was previously observed, indicating a region where a lower accuracy prediction is more probable. The ability to produce larger uncertainty estimates for cases where the model is more likely to perform poorly is desirable in the application of ship motion forecasting where conservative estimates of motions based on model uncertainty can be employed to inform decisions.
    
  \clearpage
	\subsection*{Framework}
	
	Previous methodologies of modeling ship motions with neural networks have focused on simpler application such as a 2-D midship section \citep{Xu2021, Silva2021marine, Silva2021stab}, constant speed and heading \citep{Ferrandis2021}, or extremely short time windows \citep{DAgostino2021}. However, some consideration must be made to develop a methodology to construct a generalized model capable of predicting 6-DoF motions of a free-running vessel in a high sea state. The previous work of \cite{Xu2021, Silva2021marine, Silva2021stab} was able to represent the heave and roll of a 2-D midship section that was constrained in surge, sway, and yaw by developing a relationship between the wave elevation at the inlet of the URANS computational domain and the motions of the hull. However, now that the ship is moving in space, static wave probes are not sufficient as input into the neural network, because not only must the model learn the resulting ship motions, but it also must be trained to understand the wave propagation from a static wave probe to a moving vessel with unknown trajectory. The main idea behind the present methodology, is that a vessel experiences wave excitation in the encounter frame of reference. Therefore, instantaneous wave elevation around the hull should serve as the input into the model. However, the trajectory of the vessel and thus the instantaneous encounter frame is not known \emph{a priori} and thus, must be estimated. 
	
	The following procedure describes the proposed modeling approach and training process:
	\begin{enumerate}
			\item Select $K$ wave probe locations in the initial coordinate frame
			\item Estimate the encounter frame with the surge and sway motions from the training data
			\item Find the instantaneous wave elevation at each wave probe moving with the estimated encounter frame
			\item Standardize the datasets for each respective wave probe and ship motion DoF
			\item Train the model to develop a relationship between the wave elevation time histories of the moving wave probes and the 6-DoF motions of the vessel
	\end{enumerate}

\begin{figure}[H]
		\centering
		\includegraphics[width=1\linewidth]{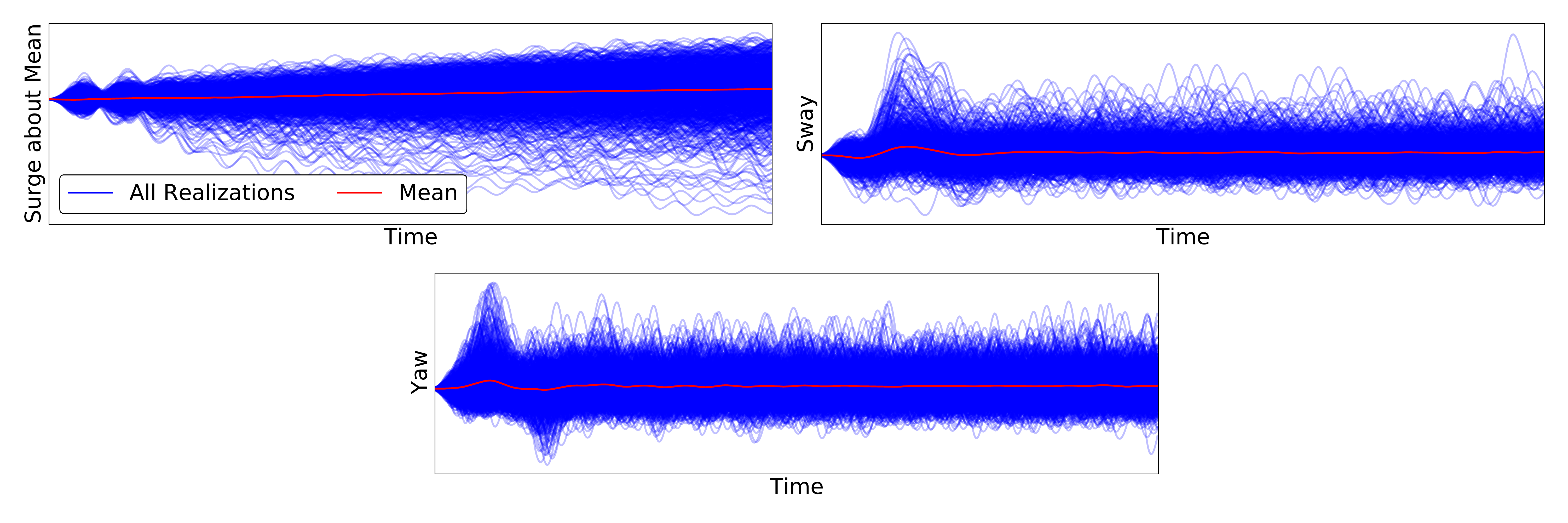}  
		\caption{Estimation of encounter frame during course-keeping based on training data.}
		\label{fig:estimateframeCourseKeeping}
	\end{figure}

	The first step in the training procedure is to select $K$ wave probe locations in the initial coordinate frame around the hull. The probe locations should be somewhat close to the hull (e.g. within a wavelength). Then, the encounter frame is estimated from surge, sway, and yaw motions in the training dataset. Fig.~\ref{fig:estimateframeCourseKeeping} shows an example of an estimated encounter frame developed from a series of training data trajectories from course-keeping simulations. The encounter frame is estimated by averaging the position of the vessel at every time step thus accounting for any drift from the mean that is common across all realizations. 
	
\begin{figure}[H]
		\centering
		\includegraphics[width=1\linewidth]{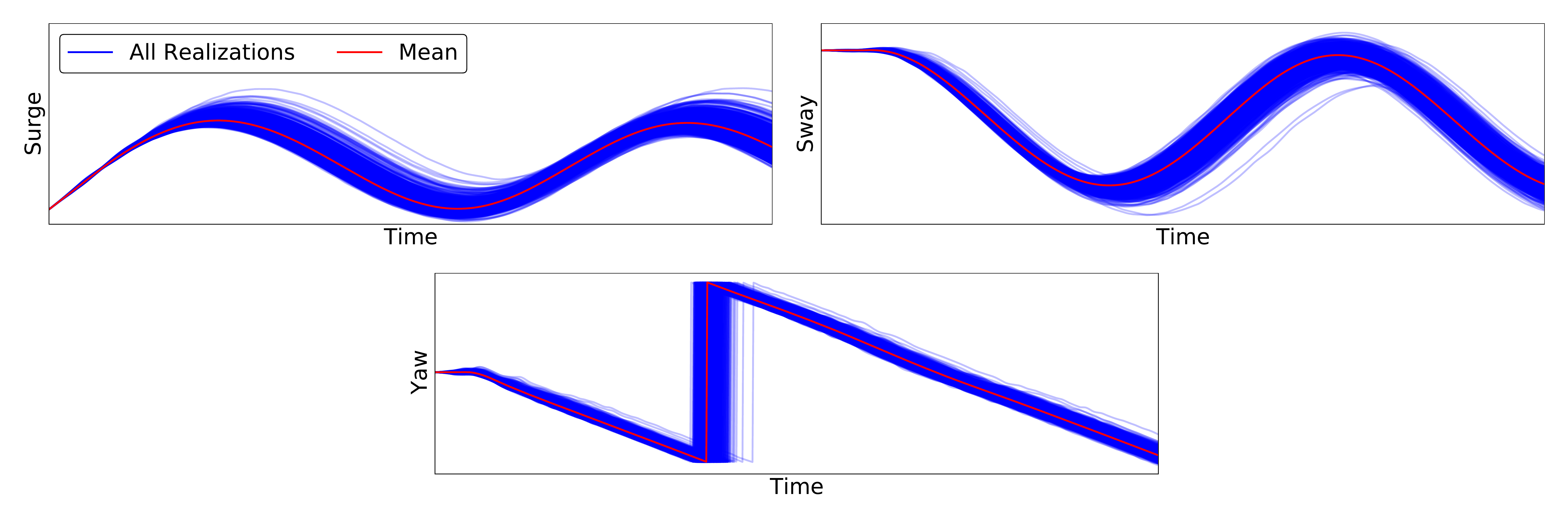}  
		\caption{Estimation of encounter frame during a turning circle based on training data.}
		\label{fig:estimateframeTurning}
	\end{figure}

	The encounter frame estimation can also be applied to general maneuvers as well like zig-zags or turning circles like shown in Fig.~\ref{fig:estimateframeTurning}. The error associated with estimating the encounter frame will be quantified later in the case study portion of the present paper where models will be trained and validated with both the estimated and actual encounter frames.
	
   Once the wave probes at the initial time step are selected and an encounter frame has been estimated, the instantaneous wave elevation at each wave probe as it travels with the encounter frame is found with Eqn.~(\ref{eq:eta}), where $\eta_k$ is the instantaneous wave elevation for a particular wave probe $k$, $\mathbf{x}_{E}$ is the coordinate location of the estimated encounter frame with respect to time $t$, $\mathbf{R}_{E}$ is the rotation matrix that describes the rotation of the encounter frame relative to the initial coordinate system, $\mathbf{x}_{k}$ is the coordinate location of wave probe $k$ at the initial time step, $a_n$, $\omega_n$, and $\phi_n$ are the amplitude, frequency, and phase, respectively, of the Fourier component $n$, and $\mathbf{k}_n$ is the wavenumber specified as a vector to denote direction of the wave components. The Fourier amplitude and frequency vectors in Eqn.~(\ref{eq:eta}) are derived from discretization of a wave spectrum. The phase vector is either random,  or determined from the reconstruction of a continuous time signal. The current implementation of the framework requires a description of wave elevation with respect to different locations in the horizontal plane and time.
   
   \begin{equation}
       \eta_k \left( \mathbf{x}_{E}, \mathbf{R}_{E}, t \right) \ = \ \sum_{n}^{} a_n cos \left( \omega_n t \ - \ \mathbf{k}_n \cdot       \left( \mathbf{x}_{E}(t) + \mathbf{R}_{E}(t) \mathbf{x}_{k}\right) \ + \ \phi_n \right)
      \label{eq:eta}
   \end{equation}
   
   The rotation matrix $\mathbf{R}_{E}(t)$ in Eqn.~(\ref{eq:R}) only considers the estimated yaw motion $\psi_E$, and is directly applied to the probe location at the initial time step when calculating the encounter frame and the resulting wave elevation.
   
   \begin{equation}
   	\mathbf{R}_{E}(t) = \left[\begin{array}{cc}
	\cos{\psi_E(t)} & -\sin{\psi_E(t)}   \\
	\sin{\psi_E(t)} &  \cos{\psi_E(t)}   \\
    \end{array}\right]
	\label{eq:R}
    \end{equation}

   Calculation of the instantaneous wave elevation at each probe moving in the estimated encounter frame builds the input training matrix $X$ in Eqn.~(\ref{eq:inputmatrix}), where each entry corresponds to a set of wave elevation values for $K$ probes. Eqn.~(\ref{eq:inputvar}) shows each entry in the $X$ training matrix for training run index $m$ ranging from 1 to $M$ and time index $t$ ranging from 1 to $T$.
   
  \begin{equation}
  X = \left[\begin{array}{cccc}
  x_{11} & x_{12} & \cdots & x_{1T} \\
  x_{21} & x_{22} & \cdots & x_{2T} \\
  \vdots & \vdots & \vdots & \vdots \\
  x_{M1} & x_{M2} & \cdots & x_{MT} \\
  \end{array}\right]
   \label{eq:inputmatrix}
  \end{equation}
  
   \begin{equation}
   x_{mt} = \left[ x_{mt}^{(1)}, \ x_{mt}^{(2)}, \ ..., \ x_{mt}^{(K)} \right]
   \label{eq:inputvar}
   \end{equation}
   
   The output training matrix is built in similar fashion with Eqn.~(\ref{eq:outputmatrix}), where each entry is the vessel's 6-DoF response in Eqn.~(\ref{eq:outputtvar}), where the entries correspond to the surge velocity, sway velocity, heave, roll, pitch, and yaw respectively for a course-keeping model and surge velocity, sway velocity, heave, roll, pitch, and yaw rate for a turning circle model. The surge and sway velocities are chosen as output for course-keeping and the surge and sway velocities and yaw rate are chosen as output for turning circles as the structure of their temporal response is more suitable for the current LSTM methodology and the trajectories could be predicted by integrating the velocity predictions. The temporal responses of the displacement of these quantities were observed to be largely non-stationary in the mean in the development of the present paper, and demonstrated poor model generalization during training. 
 
    \begin{equation}
        y = \left[\begin{array}{cccc}
        y_{11} & y_{12} & \cdots & y_{1T} \\
        y_{21} & y_{22} & \cdots & y_{2T} \\
        \vdots & \vdots & \vdots & \vdots \\
        y_{M1} & y_{M2} & \cdots & y_{MT} \\
        \end{array}\right]
        \label{eq:outputmatrix}
    \end{equation}
  
    \begin{equation}
        y_{mt} = \left[ y_{mt}^{(1)}, \ y_{mt}^{(2)}, \ y_{mt}^{(3)}, \ y_{mt}^{(4)}, \ y_{mt}^{(5)},\ y_{mt}^{(6)}\right]
        \label{eq:outputtvar}
    \end{equation}

    Once the input and output training matrices are constructed, each are standardized with respect to each input feature (i.e. individual wave probe) and output label (i.e. a single motion DoF). The standardization results in each feature and label entering the training phase with zero mean and a standard deviation of one. After the data has been standardized, the training process begins where the various parameters shown in Eqn.~(\ref{eq:f_t}) through (\ref{eq:h_t}) are continuously updated through several iterations within an optimization scheme that aims to minimize the loss function shown in Eqn.~(\ref{eq:loss}), where $\hat{y}$ is the prediction of the model and $y$ is the true output label.

    \begin{equation}
    	L(\hat{y}, y) = \frac{1}{T}\sum_{t=1}^T (\hat{y}_t - y_t)^2
    	\label{eq:loss}
    \end{equation}

	\section*{Case Studies}

\begin{figure}[H]
		\centering
		\includegraphics[width=0.9\textwidth]{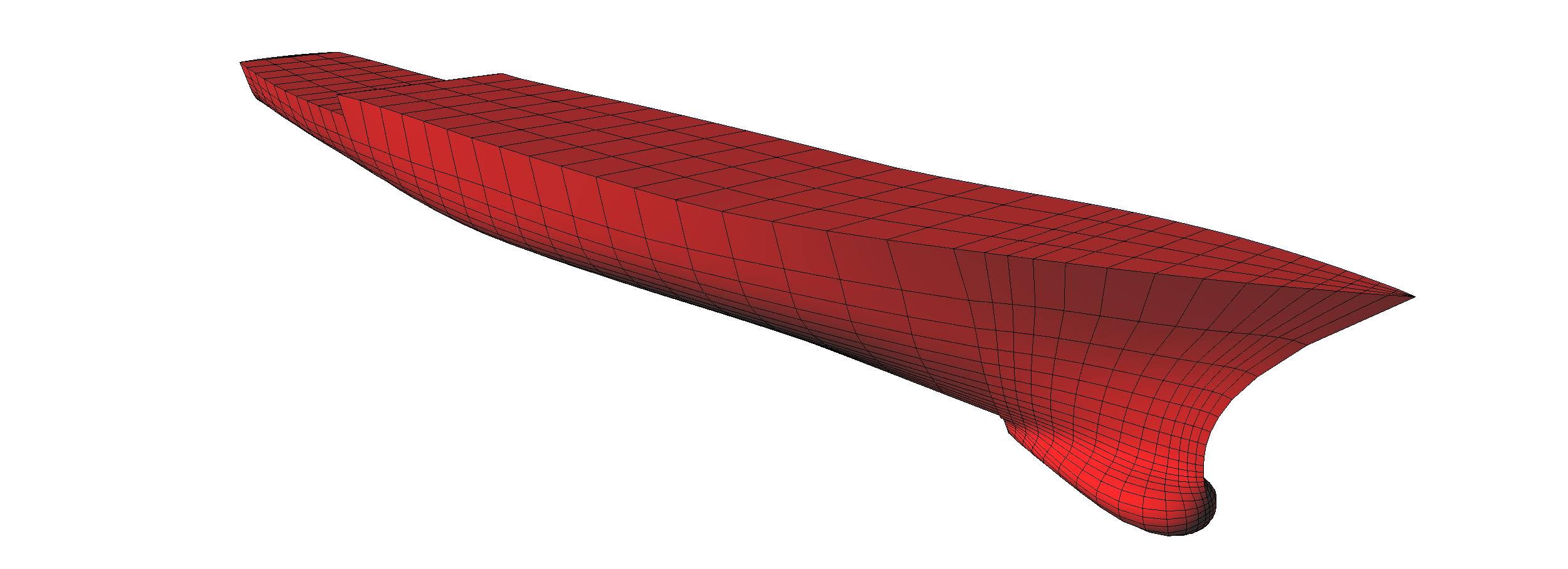}
		\caption{LAMP representation of the DTMB 5415 hullform.}
		\label{fig:LampGeo}
	\end{figure}

	The proposed methodology for representing 6-DoF ship responses with an LSTM neural network is demonstrated with simulations performed with LAMP for the DTMB 5415 hullform shown in Fig.~\ref{fig:LampGeo} for both course-keeping and turning circles. LAMP is a potential flow time-domain ship motion and wave loads simulation tool. LAMP uses a time-stepping scheme where all forces and moments acting the ship (e.g. wave-body interaction, appendages, control systems, green-water-on-deck, etc.) are evaluated at teach time step and the 6-DoF equations of motion are integrated in time to advance the solution. At the center of the LAMP calculation scheme is the solution of the three-dimensional (3-D) wave-body interaction problem \citep{Lin1990,Lin1994}. A 3-D perturbation velocity potential is computed by solving an initial boundary value problem with a potential flow boundary element method and then Bernoulli’s equation to compute the hull pressure distribution including the second-order terms. The current case study employed the LAMP-3 approach where the perturbation velocity potential is solved over the mean wetted surface (body-linear) while the nonlinear Froude-Krylov and hydrostatic restoring forces are solved over the instantaneously wetted area of the hull below the incident wave (body-nonlinear). This blended nonlinear methodology captures the significant portion of nonlinear effects in most ship-wave problems at a fraction of the computation effort for the general body-nonlinear formulation and allows for large lateral motions and generalized 6-DoF motions.
	
    Table~\ref{tab:load} shows the loading condition and fluid properties for the DTMB 5415 case study. The loading condition is derived from CFD validation studies performed for the 5415M in \cite{Sadat-Hosseini2015}, while the fluid properties represent seawater at 20 $^{\circ}$C \citep{ITTC2011}.

	\begin{table}[H]
		\caption{Loading condition of DTMB 5415 hull and fluid properties.}
		\begin{center}
			\label{tab:load}
			\begin{tabular}{l | l | l}
				\hline
				Properties & Units & Value \\
				\hline
				Length Between Perpendiculars, $L_{pp}$		&	m	&	142.0	\\
				Beam, $B$			&	m	&	19.06	\\
				Draft, $T$		&	m	&	6.15	\\
				Displacement, $\nabla$		&	tonnes	&	8431.8	\\
				Longitudinal Center of Gravity, $LCG$ (+Fwd of AP)	&	m	&	70.317	\\
				Vertical Center of Gravity, $VCG$ (ABL)	&	m	&	7.51	\\
				Transverse Metacentric Height, $GMT$		&	m	&	1.95	\\	
				Roll Gyradius, $k_{xx}$	&	m	&	7.62	\\
				Pitch Gyradius, $k_{yy}$	&	m	&	35.50	\\
				Yaw Gyradius, $k_{zz}$	&	m	&	35.50	\\
				Density of Water, $\rho_w$	&	kg/m$^\text{3}$	&	1024.81	\\	
				Kinematic Viscosity of Water, $\nu_w$	&	m$^\text{2}$/s	&	1.0508e-06	\\	
				Acceleration due to Gravity, $g$	&	m/s$^\text{2}$	&	9.80665	\\	
				\hline	
			\end{tabular}
		\end{center}
	\end{table}

    Table~\ref{tab:conditions} demonstrates a summary of the operating and seaway conditions considered for the case study as well as other details of the simulation dataset. The DTMB 5415 hull is set to operate at 20~kn in Sea State 7 stern-quartering seas described by the Bretschneider spectrum \citep{Bretschneider1959} shown in Eqn.~(\ref{eq:bretschneider}), where $\omega$ and $\omega_p$ are the wave frequency and peak wave frequency respectively. 
    
    \begin{equation}
    	S\left( \omega \right) = \frac{1.25}{4} \ \frac{\omega_p ^{4}}{\omega^{5}} \ H_s^2 \ \exp\left[-1.25 \  \left(\frac{\omega_p}{\omega}\right)^4 \right]
    	\label{eq:bretschneider}
    \end{equation}
    
    The significant wave height $H_s$, corresponds to the middle of the Sea State 7 $H_s$ band, while the peak modal period $T_p$, is the most probable modal period for Sea State 7 in the North Atlantic \citep{Bales1983}. The complete training dataset consists of 640~simulations, each 360~s in length, for a total exposure window of 64~hours. Several models are constructed in the present paper and vary in the quantity of data in the training, but they all draw from the same collection of 640~simulations. Additionally, all the models were evaluated with the same validation dataset that was independent of training dataset and contained 1000~simulations, corresponding to 100~hours of total exposure time.

	\begin{table}[H]
		\caption{Operating and seaway conditions for the DTMB 5415 case study.}
		\begin{center}
			\label{tab:conditions}
			\begin{tabular}{l | l | l}
				\hline
				Properties & Units & Value \\
				\hline
				Speed		&	knots	&	20	\\
				Wave Heading		&	deg	&	135 (Starboard Stern Quartering)	\\
				Sea State			&	-	&	7	\\
				Spectrum			&	-	&	Bretschneider	\\
				Significant Wave Height, $H_s$			&	m	&	7.5	\\
				Peak Modal Period, $T_p$			&	s	&	15 	\\
				
				Proportional Gain, $G_p$			&	-	&	4	\\
				Integral Gain, $G_i$			&	-	&	0 	\\
				Differential Gain, $G_d$			&	-	&	1 	\\
				Max Rudder Rate, $\dot{\delta}_{\rm max}$			&	deg/s	&	35 	\\
				
				Rudder Deflection (Turning Circle), $\delta_{\rm turn}$			&	deg	&	35 	\\
				Individual Run Length  & s & 360 \\
				Time Step  & s & 0.5 \\
				Number of Training Runs & - & 640 \\
				Number of Validation Runs & - & 1000 \\
				\hline	
			\end{tabular}
		\end{center}
	\end{table}

    The DTMB 5415 hull is free to surge, sway, heave, roll, pitch and yaw. The LAMP simulations utilize a quasi-steady propeller performance model from \cite{Lee2003} for both the course-keeping and turning circle cases. Course-keeping is maintained with two rudders modeled as low aspect ratio foils controlled by the PID controller shown in Eqn.~(\ref{eq:pid}). The commanded rudder deflection $\delta_c$ is calculated with the proportional $G_p$, integral $G_i$, and derivative $G_d$ gains and the desired heading $\psi_d$ and the current heading $\psi$. While, the turning circle cases maintain a 35~deg rudder deflection during the entire simulation.
    
    \begin{equation}
	\delta_c = G_p(\psi_d - \psi) + G_i\int_{0}^{t} (\psi_d - \psi(\tau)) d\tau + G_d\dot{\psi}
	\label{eq:pid}
    \end{equation}

    To evaluate the efficacy of the proposed LSTM neural network model building methodology, understanding the behavior and convergence is imperative for models trained with different quantities of data and the fidelity of the wave field description. Table~\ref{tab:trainmatrix} shows the different neural network parameters explored in the training of the models. The number of training runs varies from 10~to~640, while the number of wave probes for the input ranged from 1~to~27, for a total of 28~unique models. The same neural network architecture, hyper-parameters, and training approach are implemented for each of the constructed models. The subsequent validation of the different models is based on the same set of validation runs, unseen by all of the developed models during training. All of the neural network models are implemented with the Keras \citep{Chollet2015}, a high-level Application Programming Interface (API) for Tensorflow \citep{Abadi2015}.
    
	\begin{table}[H]
		\caption{Training matrix, neural network architecture, and hyper-parameters for the DTMB 5415 case study.}
		\begin{center}
			\label{tab:trainmatrix}
			\begin{tabular}{l | l | l}
				\hline
				Properties & Units & Value \\
				\hline
				Number of Training Runs		&	-	&	10, 20, 40, 80, 160, 320, 640	\\
				Number of Wave Probes		&	-	&	1, 3, 9, 27	\\
				Number of Time Steps per Run		&	-	&	720	\\
				Number of Units per Layer		&	-	&	250	\\
				Number of Layers		&	-	&	3	\\
				Dropout		&	-	&	0.1	\\
				Learning Rate		&	-	&	0.00001	\\
				Number of Epochs		&	-	&	2000	\\
				Optimizer		&	-	&	Adam \citep{kingma2014adam}	\\
				\hline	
			\end{tabular}
		\end{center}
	\end{table}

\begin{figure}[H]
	    \centering
	    \includegraphics[width=1\textwidth]{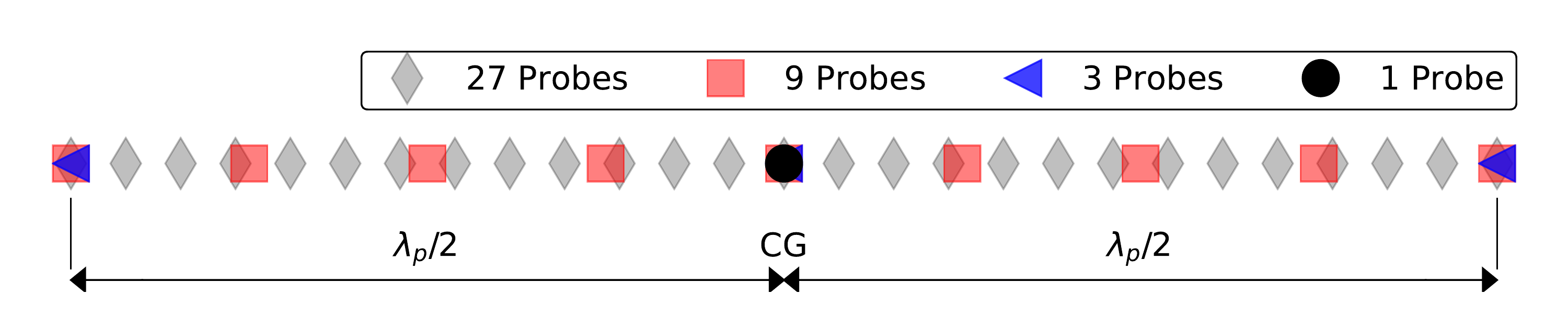}
        \caption{Probe locations relative to the ship's CG.}
	    \label{fig:probes}
    \end{figure}

    The probe locations considered for the case study are shown in Fig.~\ref{fig:probes}, where $\lambda_p$ is a nominal wavelength calculated based on $T_p$. When only one probe is considered, that wave probe is located at the center of gravity ($CG$). Models built with more than one wave probe share the same extents based on $\lambda_p$ but the level of discretization is increased between the extents is increased. In the current paper, wave probe locations are only varied longitudinally as there no improvement in model accuracy is observed when probes are placed away from the ship's centerline. This may be specific to the current case and future work will have to investigate the influence of lateral wave probe locations.

    \clearpage
    \subsection*{Course-keeping in Irregular Waves}
	
	A series of models are constructed to evaluate their accuracy and convergence with respect to the quantity of input wave probes and training data for course-keeping LAMP simulations. Following the work of \cite{Xu2021}, the $L_2$ and $L_\infty$ error between the LAMP simulations $y$ and neural network predictions $\hat{y}$ are used to evaluate the accuracy of the different models. The $L_2$ error in Eqn.~(\ref{eq:L2error}) is a measure of the mean squared error across a particular time series with time indices ranging from $i$~to~$T$, while the $L_\infty$ error in Eqn.~(\ref{eq:Linferror}) quantifies the largest error observed in a given time series.

	\begin{equation}
		L_2(y, \hat{y}) = \sqrt{\frac{1}{T}\sum_{i=1}^T (y_i - \hat{y}_i)^2}
		\label{eq:L2error}
	\end{equation}

    \begin{equation}
    	L_\infty(y, \hat{y}) = \max_{i=1, \cdots, T} |y_i - \hat{y}_i|
    	\label{eq:Linferror}
    \end{equation}

\begin{figure}[H]
		\centering
			\includegraphics[width=0.9\textwidth]{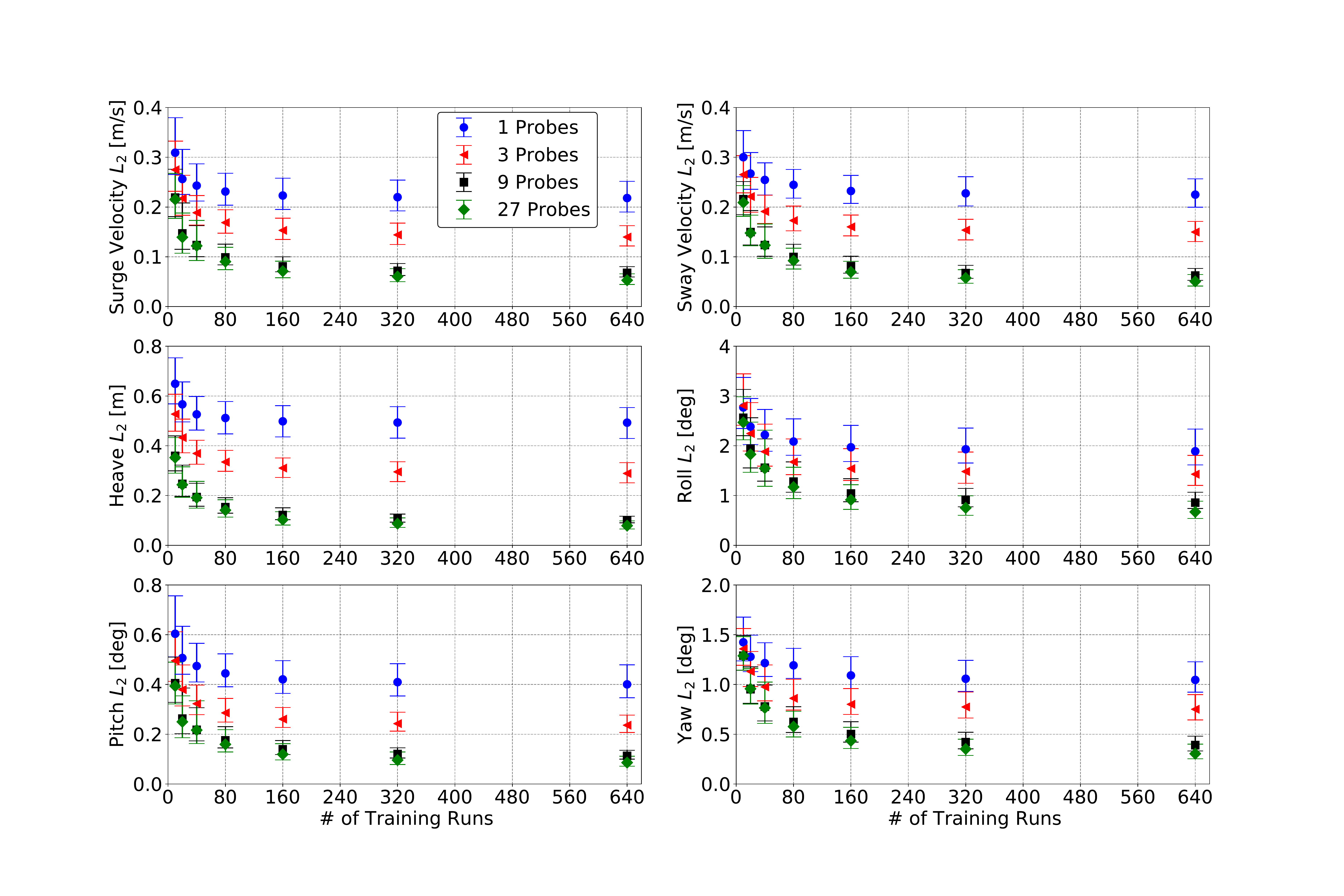}
		\caption{Comparison of $L_2$ error for each DoF in the course-keeping case study.}
		\label{fig:L2Compare_courseKeeping}
	\end{figure}

\begin{figure}[H]
		\centering
			\includegraphics[width=0.9\textwidth]{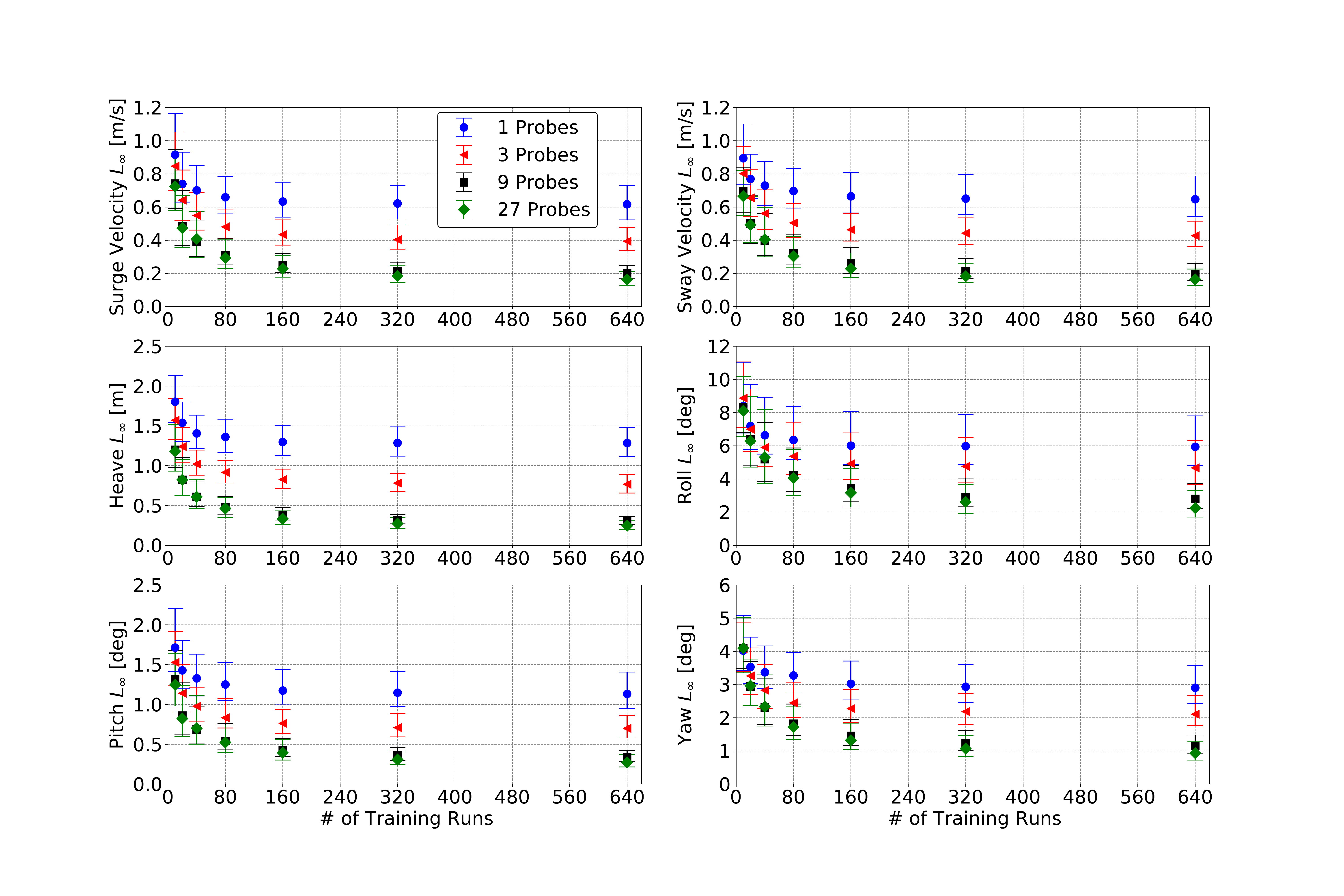}
		\caption{Comparison of $L_\infty$ error for each DoF in the course-keeping case study.}
		\label{fig:LinfCompare_courseKeeping}
	\end{figure}

    Fig.~\ref{fig:L2Compare_courseKeeping} and \ref{fig:LinfCompare_courseKeeping} show comparisons of the $L_2$ and $L_\infty$ error at each DoF for neural network models trained with various quantities of training data and wave probes as input into the models. Each marker denotes the median error for all of the runs, while the upper and lower error bars correspond to the 75\% and 25\% quantiles, respectively. The data trend towards smaller error as both the number of wave probes and training data quantity is increased. As the median error decreases, so does the spread of the total error denoted by the error bars. Though the error is decreased by increasing the number of wave probes from 9~to~27, the difference is much smaller in comparison to the coarser wave descriptions with 1 and 3 probes. Additionally, though the difference between models decreases as number of training runs increases from 320~to~640, the overall trends of the models indicate that the predictions are continuing to improve as the training data quantity increases. This convergence provides confidence in the modeling approach and indicates that in a real-time forecasting implementation onboard a vessel, the continual addition of more data will help improve the model over time.

\begin{figure}[H]
		\centering
		\includegraphics[width=1\textwidth]{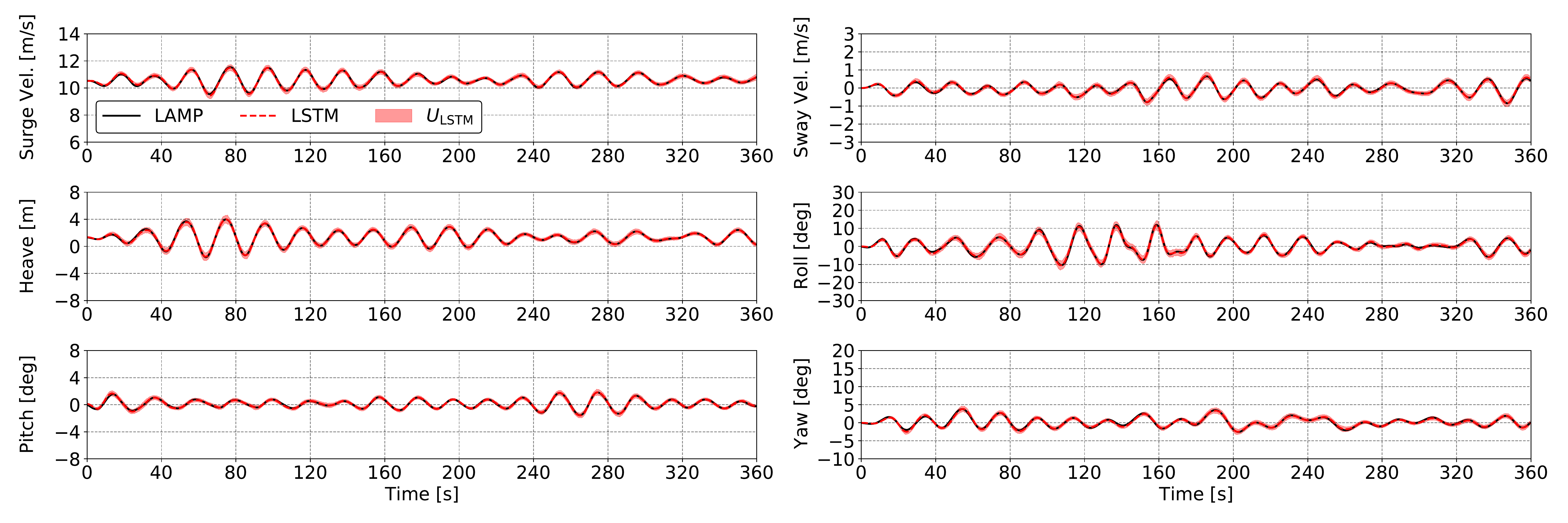}
		\caption{Best (ranked by $L_2$ error) motion predictions for a model with 27 probes and 640 training runs in the course-keeping case study.}
		\label{fig:BestL2TH_courseKeeping}
	\end{figure}

\begin{figure}[H]
		\centering
		\includegraphics[width=1\textwidth]{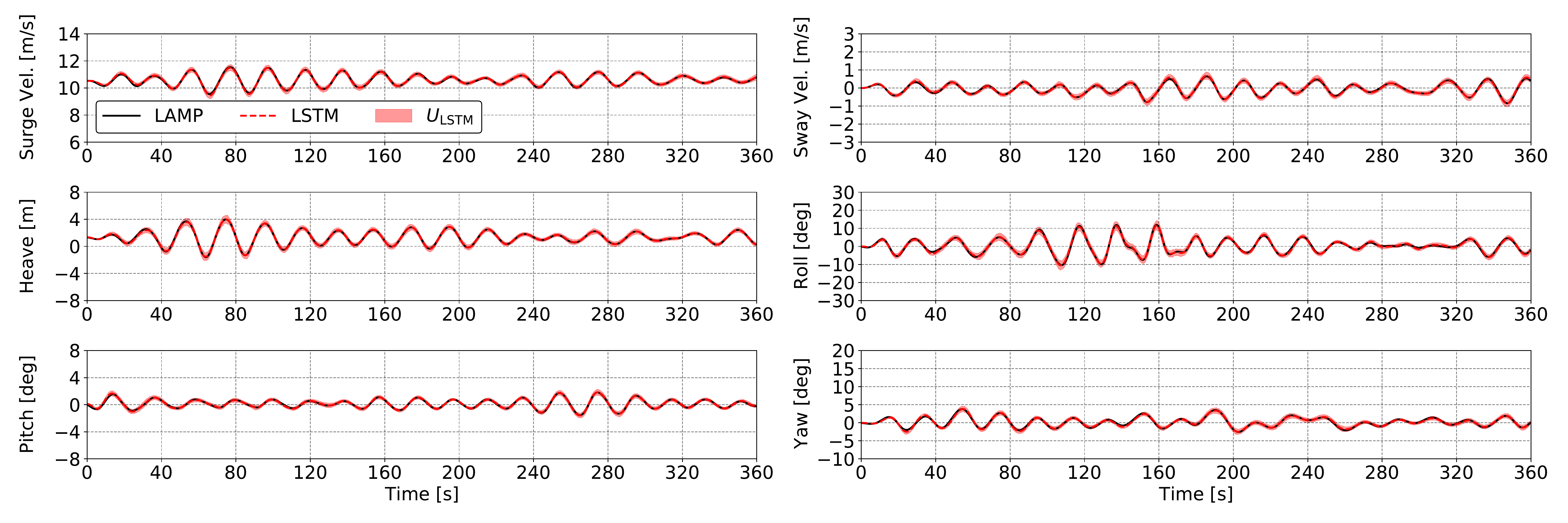}
		\caption{Best (ranked by $L_\infty$ error) motion predictions for a model with 27 probes and 640 training runs in the course-keeping case study.}
		\label{fig:BestLinfTH_courseKeeping}
	\end{figure}

    Fig.~\ref{fig:L2Compare_courseKeeping} and \ref{fig:LinfCompare_courseKeeping} demonstrate the increase in accuracy as the quantity of wave probes and training data is increased, resulting in the most accurate model being built with 27 wave probes and 640 training data runs. Though the comparison of $L_2$ and $L_\infty$ provides an overall assessment of the models, the objective of ship motion forecasting is to produce an accurate temporal response. Fig.~\ref{fig:BestL2TH_courseKeeping} and \ref{fig:BestLinfTH_courseKeeping} show comparisons between LAMP and the neural network model built with 27 wave probes and 640 training data runs for the validation runs with the lowest $L_2$ and $L_\infty$ error for each DoF. The LSTM prediction, denoted by a red dashed line, is the mean of the stochastic predictions made by the Monte Carlo dropout approach while the uncertainty $U_\mathrm{LSTM}$, highlighted in red, is calculated to be $\pm 5\sigma$ based on the Monte Carlo dropout realizations. $\pm 5\sigma$ is chosen for the uncertainty interval as the overall uncertainty in the developed models is low and increasing the interval size allowed the larger uncertainty regions to be observed visually. Overall, for the cases with the lowest $L_2$ and $L_\infty$ error, the LSTM model predicts the 6-DoF response well and the uncertainty, $U_\mathrm{LSTM}$ predicted with the Monte Carlo dropout approach is small.
    
\begin{figure}[H]
		\centering
		\includegraphics[width=1\textwidth]{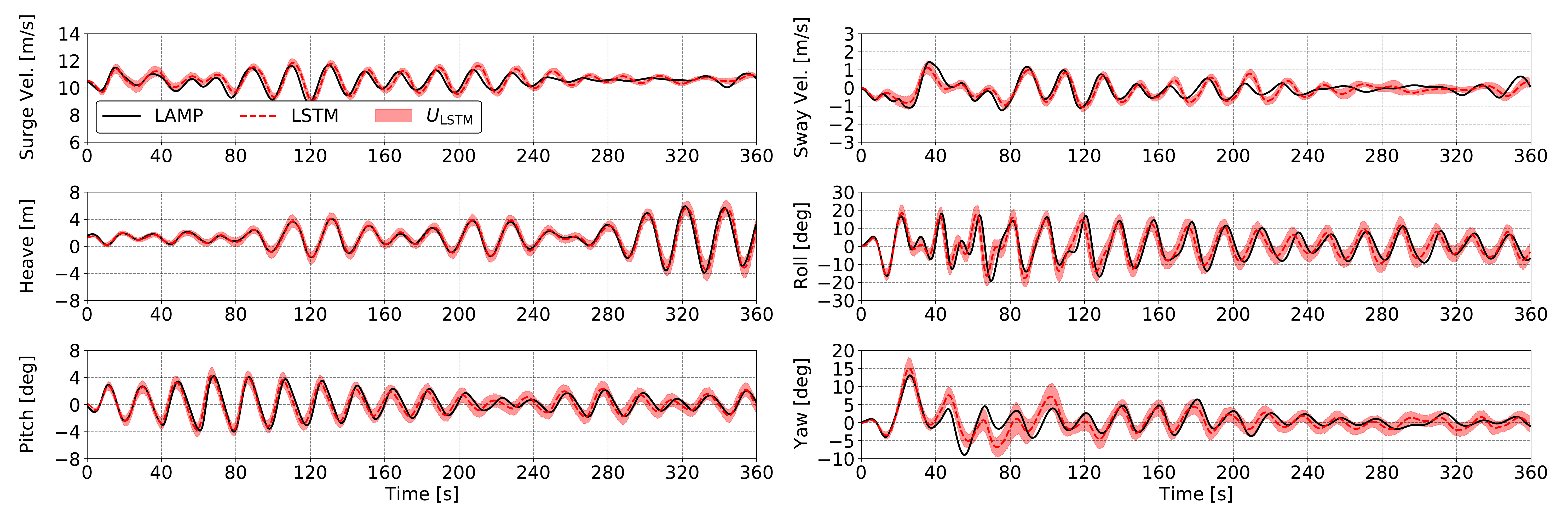}
		\caption{Worst (ranked by $L_2$ error) motion predictions for a model with 27 wave probes and 640 training runs in the course-keeping case study.}
		\label{fig:WorstL2TH_courseKeeping}
	\end{figure}

\begin{figure}[H]
		\centering
		\includegraphics[width=1\textwidth]{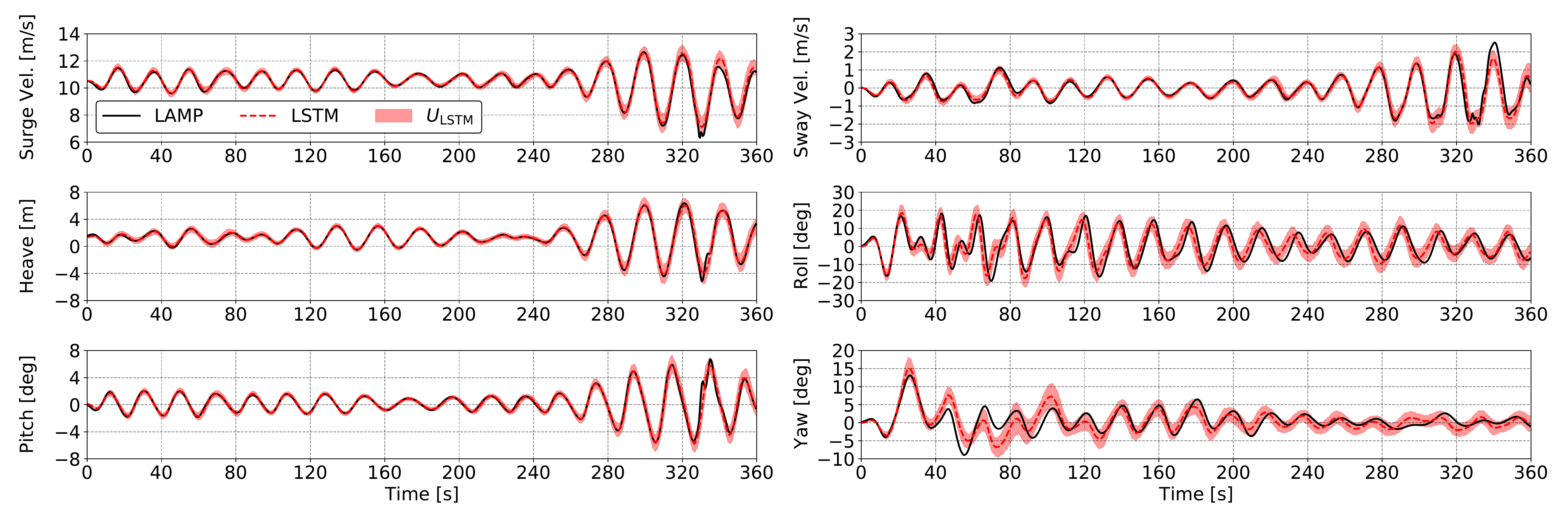}
		\caption{Worst (ranked by $L_\infty$ error) motion predictions for a model with wave 27 probes and 640 training runs in the course-keeping case study.}
		\label{fig:WorstLinfTH_courseKeeping}
	\end{figure}

     Fig.~\ref{fig:WorstL2TH_courseKeeping} and \ref{fig:WorstLinfTH_courseKeeping} show comparisons between LAMP and the LSTM model built with 27 wave probes and 640 training data runs for the cases with the largest $L_2$ and $L_\infty$ error for each DoF. The magnitude of the responses is much greater in Fig.~\ref{fig:WorstL2TH_courseKeeping} and \ref{fig:WorstLinfTH_courseKeeping} than in Fig.~\ref{fig:L2Compare_courseKeeping} and \ref{fig:LinfCompare_courseKeeping}, thus the model is capable of providing better predictions for cases with an overall smaller response. In contrast with Fig.~\ref{fig:BestL2TH_courseKeeping} and \ref{fig:BestLinfTH_courseKeeping}, the uncertainty is larger in Fig.~\ref{fig:WorstL2TH_courseKeeping} and \ref{fig:WorstLinfTH_courseKeeping}, especially for the DoF and sequences where the LSTM prediction is poor, such as the roll and yaw motions between 40-120~s. Though the predictions are not as accurate as what is shown in Fig.~\ref{fig:BestL2TH_courseKeeping} and \ref{fig:BestLinfTH_courseKeeping}, the predictions are reasonably accurate, the uncertainty estimate accurately predicts the larger error between the model and the unseen LAMP predictions.

\begin{figure}[H]
		\centering
		\includegraphics[width=0.9\textwidth]{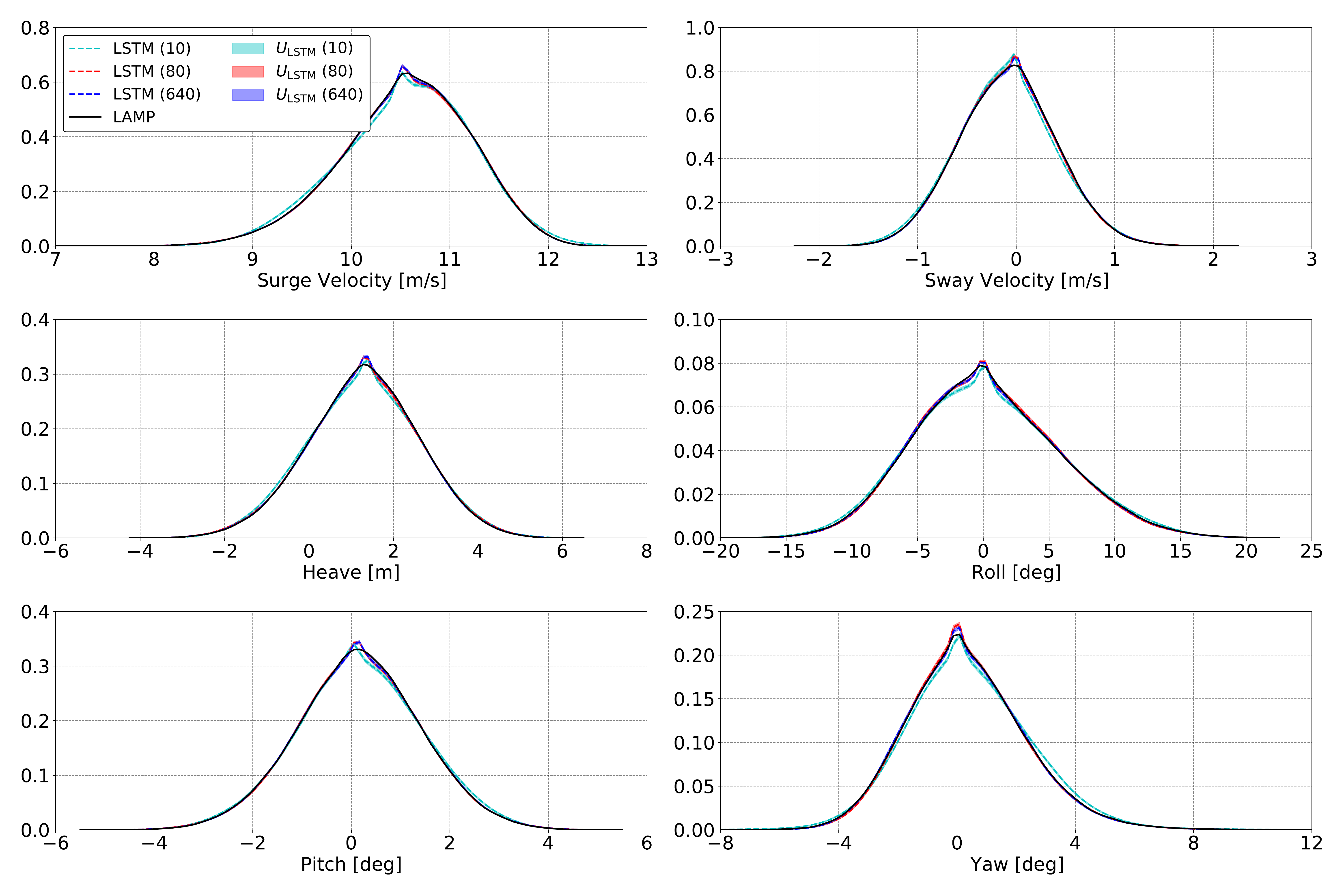}
		\caption{Comparison of the PDF for each DoF with models trained with 27 wave probes in the course-keeping case study.}
		\label{fig:compareDist_courseKeeping}
	\end{figure}

\begin{figure}[H]
		\centering
		\includegraphics[width=0.9\textwidth]{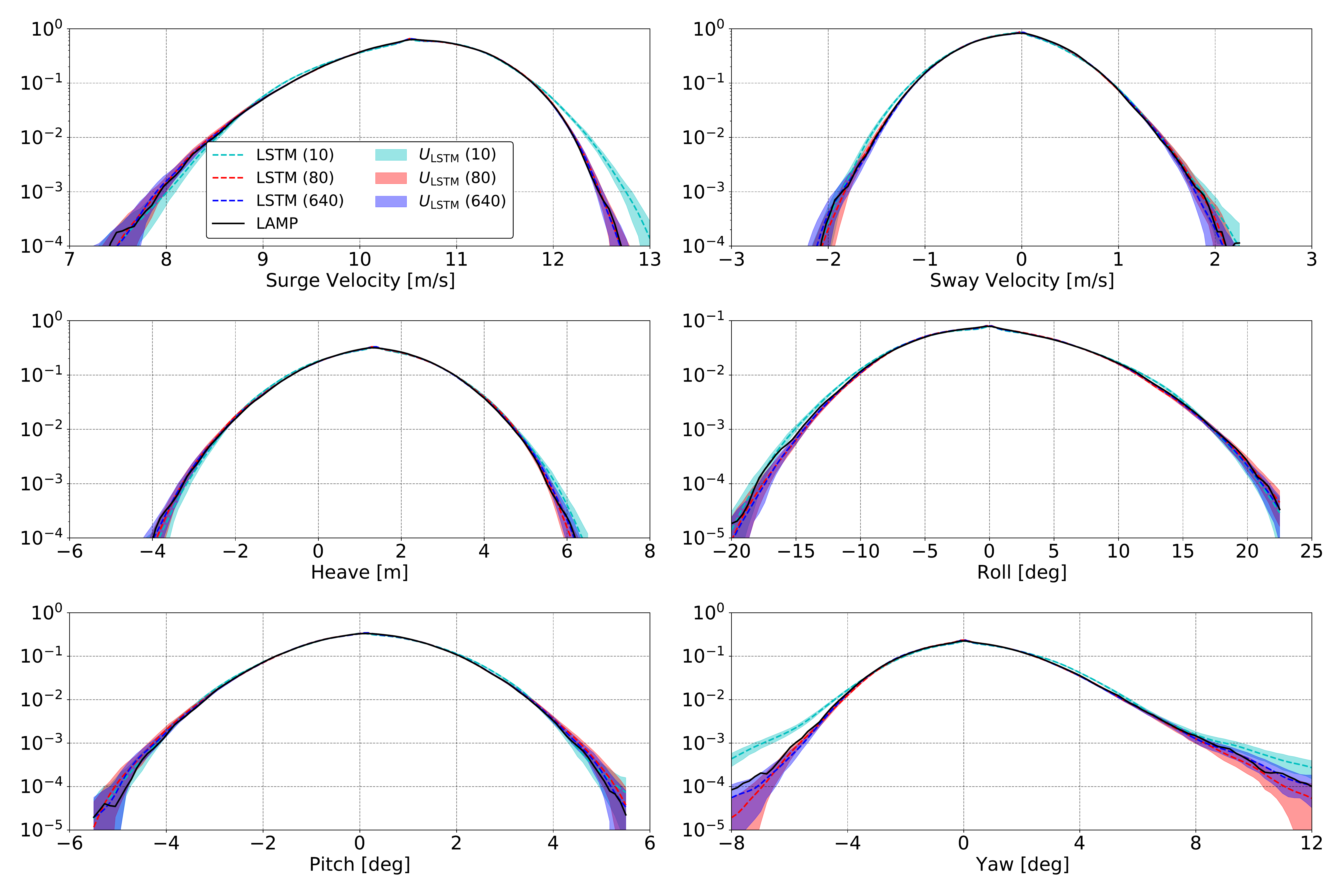}
		\caption{Comparison of the PDF tails for each DoF with models trained with 27 wave probes in the course-keeping case study.}
		\label{fig:compareDistlog_courseKeeping}
	\end{figure}

	Though temporal responses are critical for real-time ship motion forecasting, understanding the probability distribution of a particular response is desirable to produce a statistical description of the ship's operability. Comparisons of probability distributions also provide a useful method of validating models and have been leveraged in previous work validating CFD tools in \cite{Serani2021} and neural network models in \cite{Silva2021marine,Silva2021stab}. Fig.~\ref{fig:compareDist_courseKeeping} shows a comparison of the probability distribution function (PDF) for each DoF, for models built with 27 wave probes and training data of 10, 80, and 640 runs. Overall, the PDF comparisons are accurate at even 10 runs. Indicating that a neural network model trained with 10 runs can produce statistical distributions that represent those predicted by LAMP. Fig.~\ref{fig:compareDistlog_courseKeeping} shows the PDF with a logarithmic scale that emphasizes the larger response in the tails of the distributions. Even when comparing the tails, the neural network predictions do well in reproducing motions larger than the statistical mean. The model trained with 10 runs does show a larger discrepancy that is not observed Fig.~\ref{fig:compareDist_courseKeeping}, but by increasing the amount of training data, the models start to converge onto the LAMP predictions and the uncertainty in the tail region also reduces. Fig.~\ref{fig:compareDistlog_courseKeeping} indicates that this approach, with some considerations, could be useful for predicting extremes such as what was done for the CWG method in \cite{Silva2021marine,Silva2021stab} but also other extreme event probabilistic frameworks, where a fast-running surrogate model capable of producing temporal responses is desired.
    
\begin{figure}[H]
	\centering
	\includegraphics[width=0.9\textwidth]{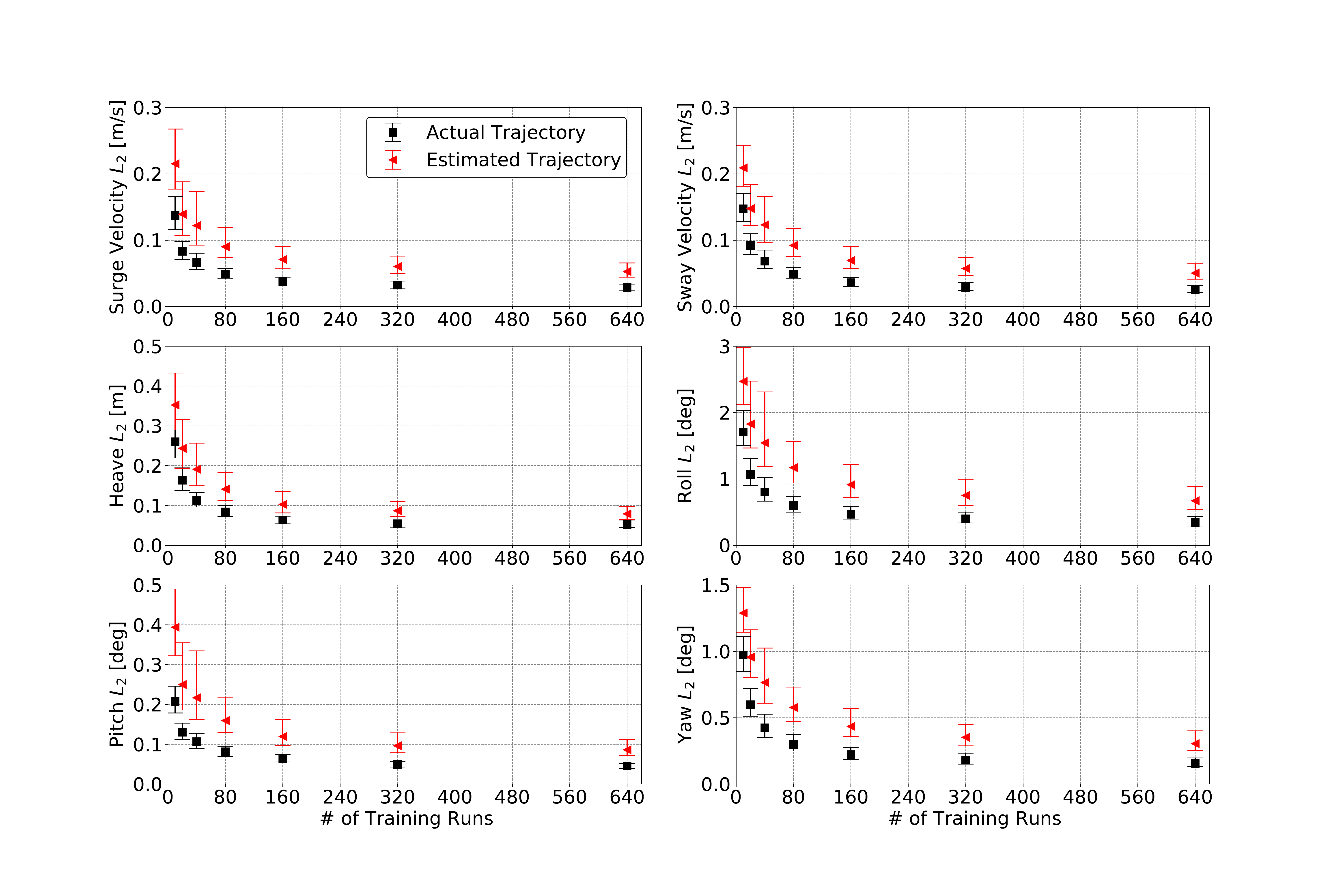}
	\caption{Comparison of $L_2$ error for each DoF with models trained with the actual and estimated encounter frames in the course-keeping case study.}
	\label{fig:L2Compare_knownTrajectory_courseKeeping}
	\end{figure}

\begin{figure}[H]
	\centering
	\includegraphics[width=0.9\textwidth]{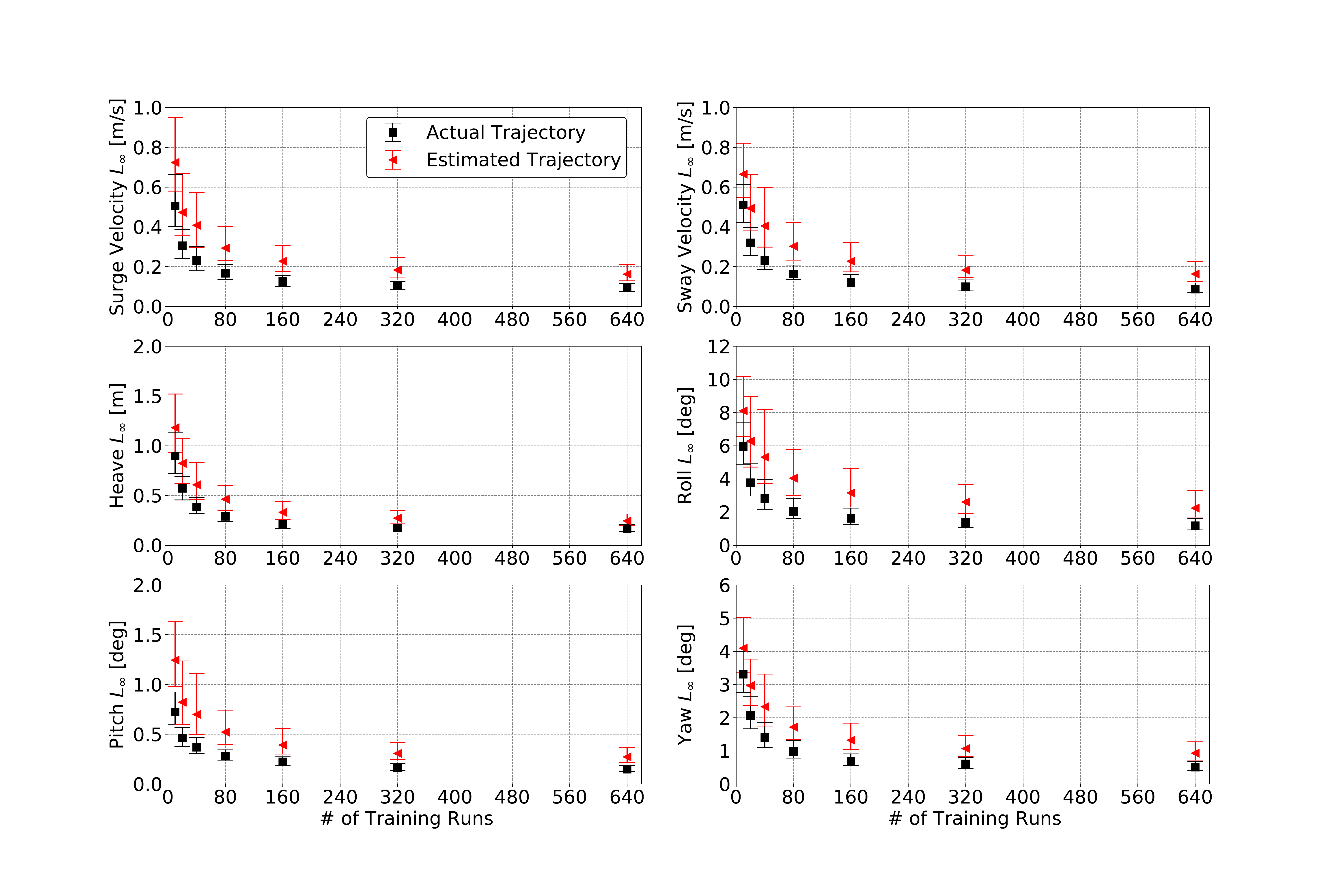}
	\caption{Comparison of $L_\infty$ error for each DoF with models trained with the actual and estimated encounter frames in the course-keeping case study.}
	\label{fig:LinfCompare_knownTrajectory_courseKeeping}
	\end{figure}

    A key limitation of the present methodology is that the encounter frame must be estimated, as the actual trajectory of the vessel in waves is not known \emph{a priori}. Therefore, the wave probes utilized as input into the neural network models are not in the actual instantaneous encounter frame of the vessel. Therefore, large enough deviations between the estimated and actual frame can produce bad predictions. To investigate the accuracy of the estimated encounter frame methodology, separate models are built with 27 wave probes in the actual encounter frame from each training run and predictions are made with validation dataset and the actual encounter frame from each of those runs as well. The difference between these new set of models and the ones built with the estimated encounter frame is a direct quantification of the consequences of estimating the encounter frame. Fig.~\ref{fig:L2Compare_knownTrajectory_courseKeeping} and \ref{fig:LinfCompare_knownTrajectory_courseKeeping} show the comparison between models with the actual and estimated encounter frames for both the $L_2$ and $L_\infty$ error for each DoF. By utilizing the actual encounter frame in the training and inference of model, the error is roughly half of the estimated frame. These comparisons indicate that a better estimate of the encounter frame could further reduce the error without requiring more training data. Future work should explore an encounter frame estimator that is wave excitation-dependent to further improve the models presented.
    
	\subsection*{Turning Circle in Irregular Waves}
	
\begin{figure}[H]
		\centering
		\includegraphics[width=0.9\textwidth]{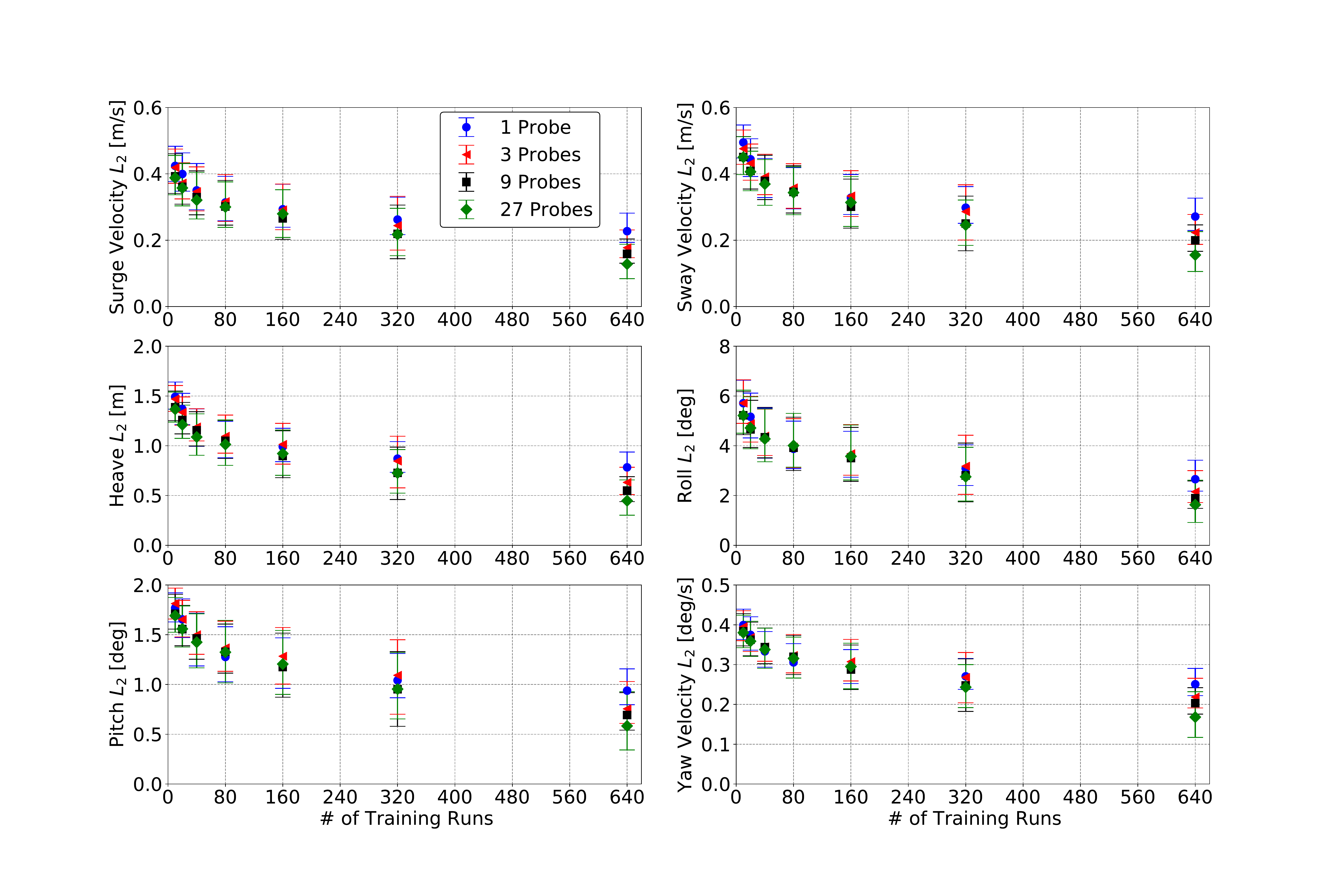}
		\caption{Comparison of $L_2$ error for each DoF in the turning circle case study.}
		\label{fig:L2Compare_turningCircle}
	\end{figure}

\begin{figure}[H]
		\centering
		\includegraphics[width=0.9\textwidth]{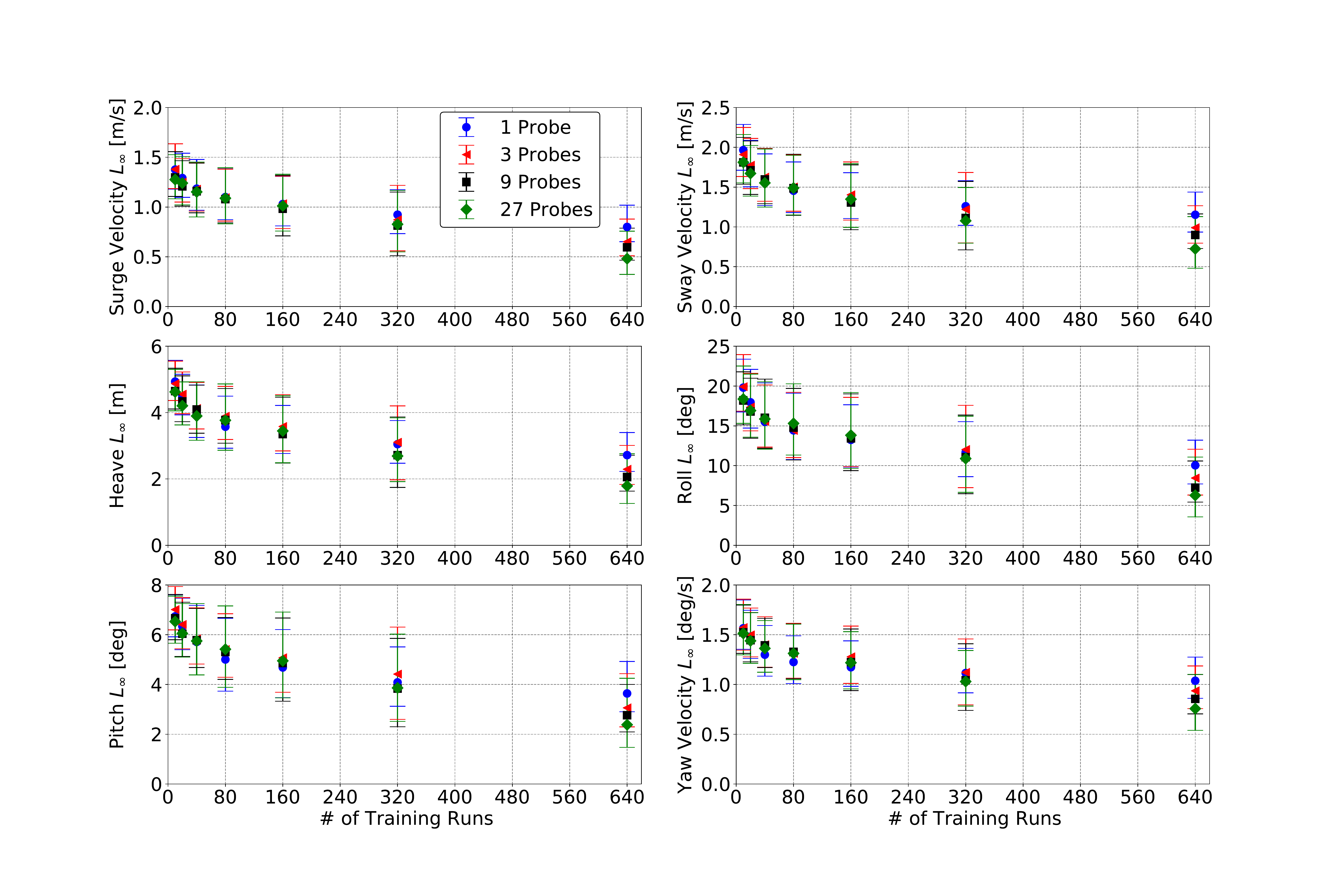}
		\caption{Comparison of $L_\infty$ error for each DoF in the turning circle case study.}
		\label{fig:LinfCompare_turningCircle}
	\end{figure}

	A separate set of models are constructed for turning circle simulations and evaluated for their accuracy and convergence with respect to the quantity of input wave probes and training data in the same manner that is performed for the course-keeping case study. The turning circle models are built in the same manner as the course-keeping models, except that the encounter frame is estimated and the models are trained and evaluated with only turning circle simulations. The turning circle models are evaluated in the same manner that was utilized for the course-keeping models. Fig.~\ref{fig:L2Compare_turningCircle} and \ref{fig:LinfCompare_turningCircle} show the comparison of $L_2$ and $L_\infty$ error for each DoF for the turning circle case study with the validation runs. The trends are similar to what is observed for course-keeping, where increasing the quantity of wave probes and training data leads to a more accurate model. However, the decrease in error as the quantity of wave probes increases is less dramatic for the turning circles and the overall error for all models is larger in comparison to the course-keeping.
	
\begin{figure}[H]
		\centering
		\includegraphics[width=1\textwidth]{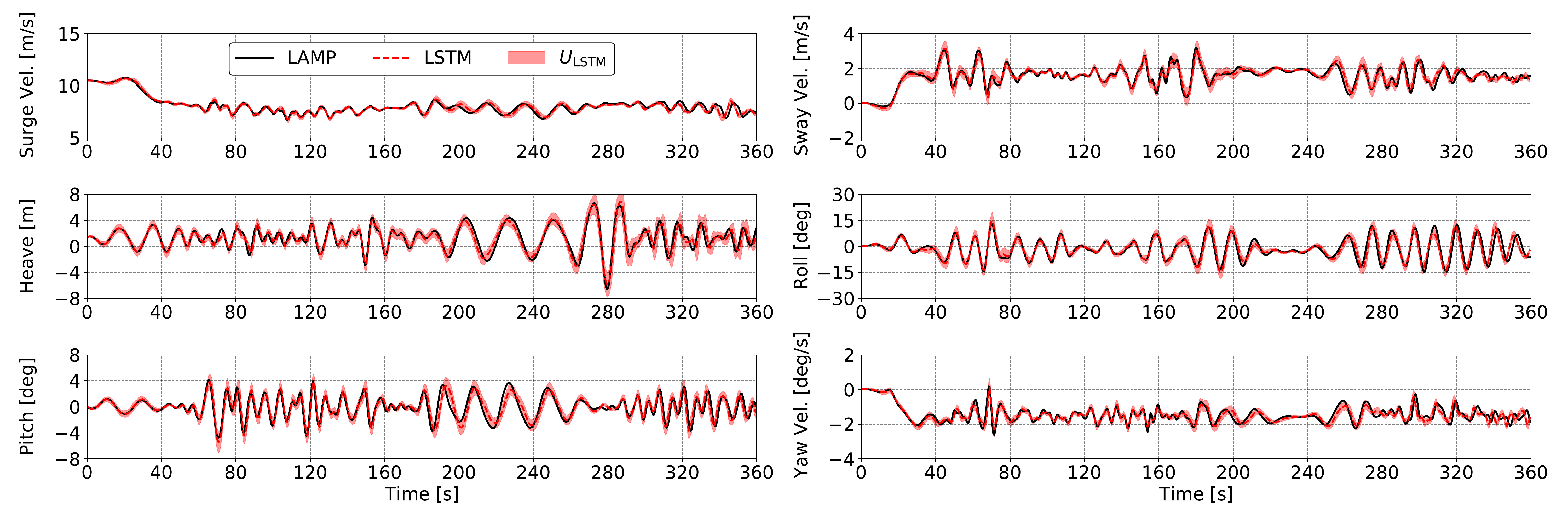}
		\caption{Best (ranked by $L_2$ error) motion predictions for a model with 27 probes and 640 training runs in the turning circle case study.}
		\label{fig:BestL2TH_turningCircle}
	\end{figure}

\begin{figure}[H]
		\centering
		\includegraphics[width=1\textwidth]{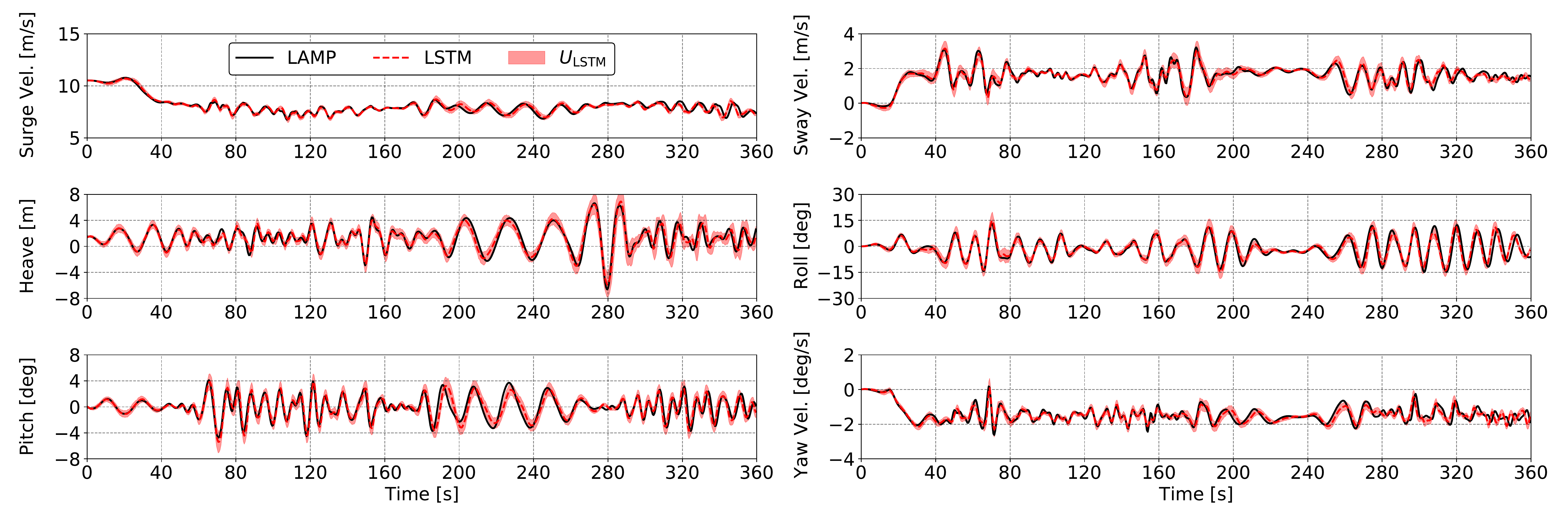}
		\caption{Best (ranked by $L_\infty$ error) motion predictions for a model with 27 probes and 640 training runs in the turning circle case study.}
		\label{fig:BestLinfTH_turningCircle}
	\end{figure}

	Similar to the course-keeping case, the most accurate model for the turning circle case is built with 27~wave probes and 640~training data runs. Fig.~\ref{fig:BestL2TH_turningCircle} and \ref{fig:BestLinfTH_turningCircle} show the temporal response comparison between LAMP and the LSTM model for the validation runs with the lowest $L_2$ and $L_\infty$ error for each DoF for the turning circle case study. Overall, the LSTM model predicts the 6-DoF response well with low uncertainty.

\begin{figure}[H]
		\centering
		\includegraphics[width=1\textwidth]{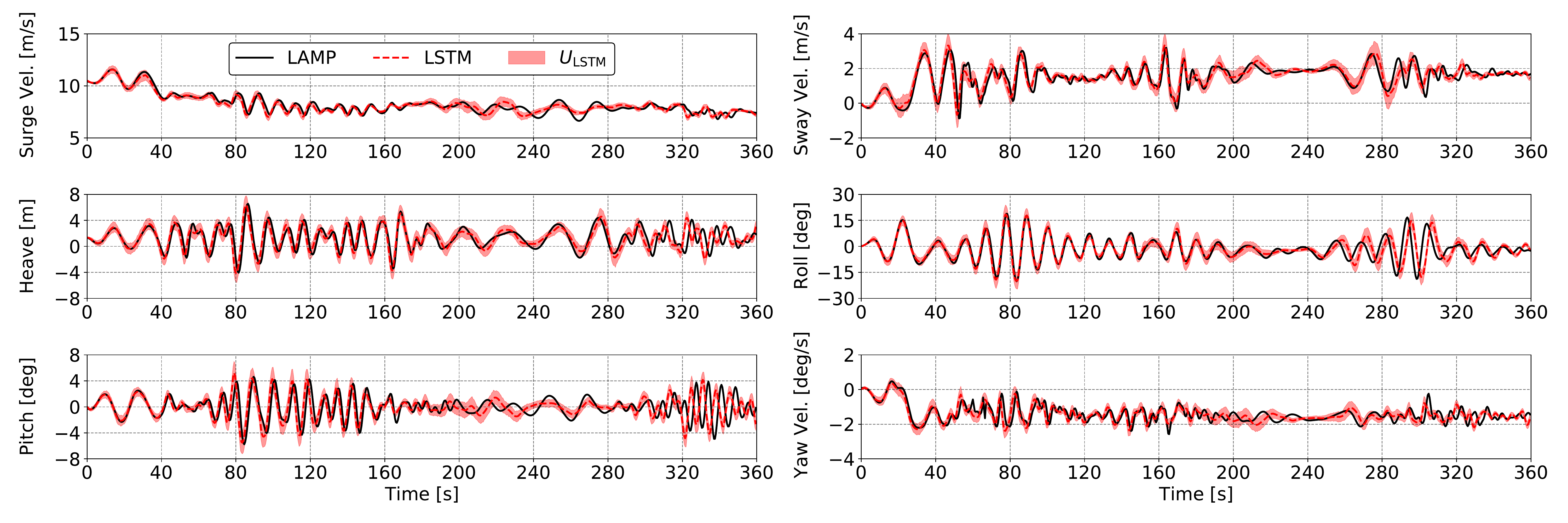}
		\caption{Worst (ranked by $L_2$ error) motion predictions for a model with 27 probes and 640 training runs in the turning circle case study.}
		\label{fig:WorstL2TH_turningCircle}
	\end{figure}

\begin{figure}[H]
		\centering
		\includegraphics[width=1\textwidth]{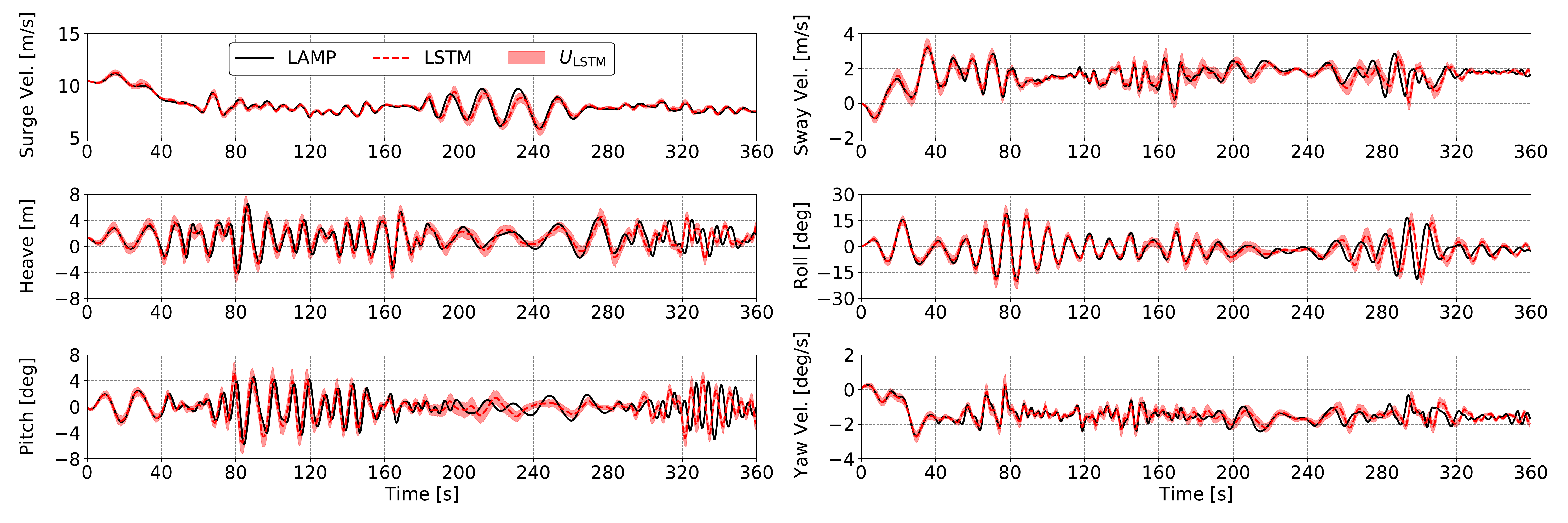}
		\caption{Worst (ranked by $L_\infty$ error) motion predictions for a model with 27 probes and 640 training runs in the turning circle case study.}
		\label{fig:WorstLinfTH_turningCircle}
	\end{figure}

	Fig.~\ref{fig:WorstL2TH_turningCircle} and \ref{fig:WorstLinfTH_turningCircle} show the comparison between LAMP and the LSTM model built with 27~wave probes and 640~training data runs for the validation runs with the largest $L_2$ and $L_\infty$ error for each DoF in the turning circle case study. Some regions of the temporal response are predicted well, while others there is a large discrepancy between LAMP and the LSTM model that is driving the error. The overall uncertainty in the time-histories is larger in those regions, which can provide an indicator of a lower accuracy prediction. The response of the vessel in the turning circle leads to a higher frequency response than what was observed in the course-keeping simulations. The difference in response frequency could be what is driving the overall error increase for the turning circle and indicates that these dynamical responses are more difficult to learn with the current methodology.

\begin{figure}[H]
		\centering
		\includegraphics[width=0.9\textwidth]{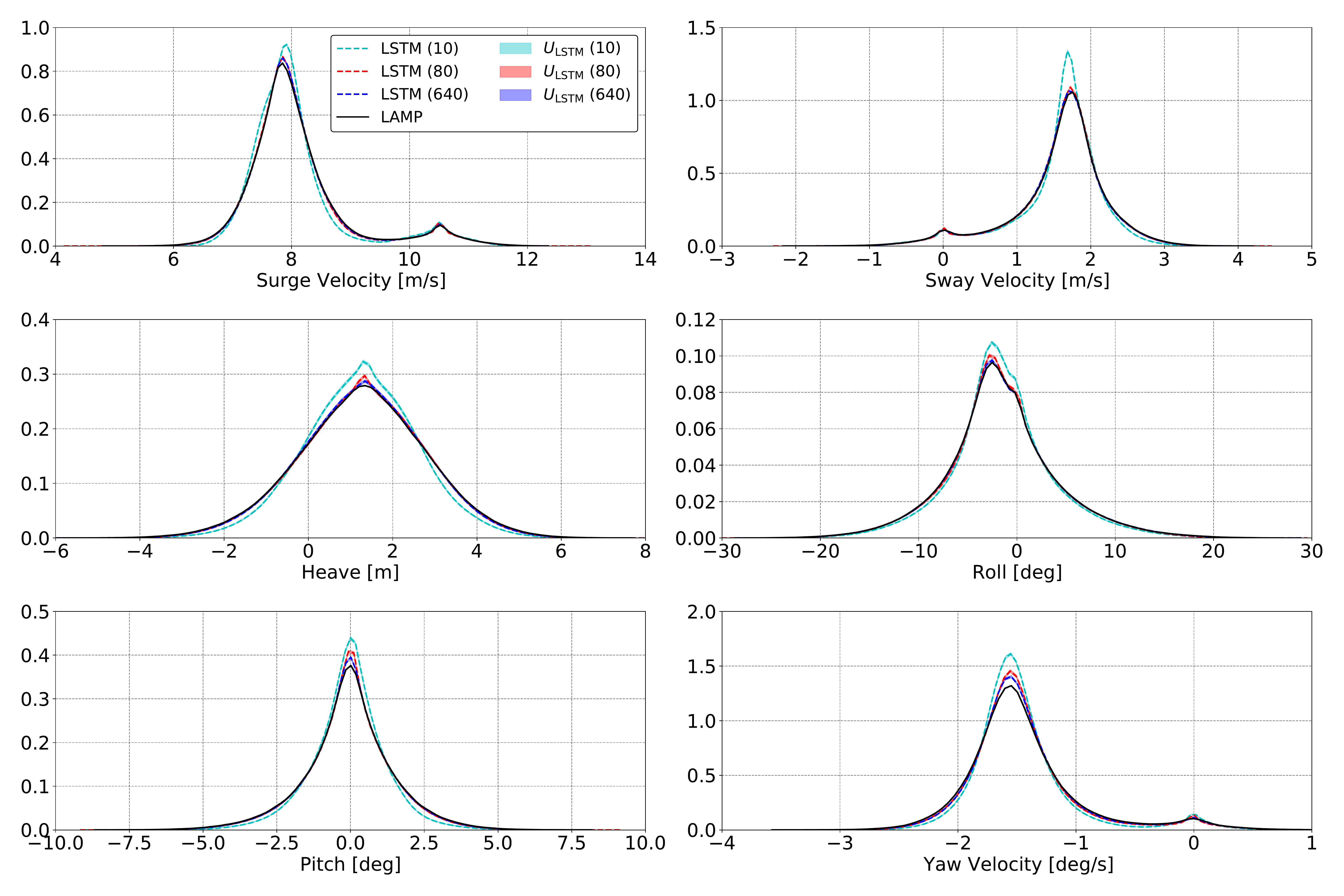}
		\caption{Comparison of the PDF for each DoF with models trained with 27 wave probes in the turning circle case study.}
		\label{fig:compareDist_turningCircle}
	\end{figure}

\begin{figure}[H]
		\centering
		\includegraphics[width=0.9\textwidth]{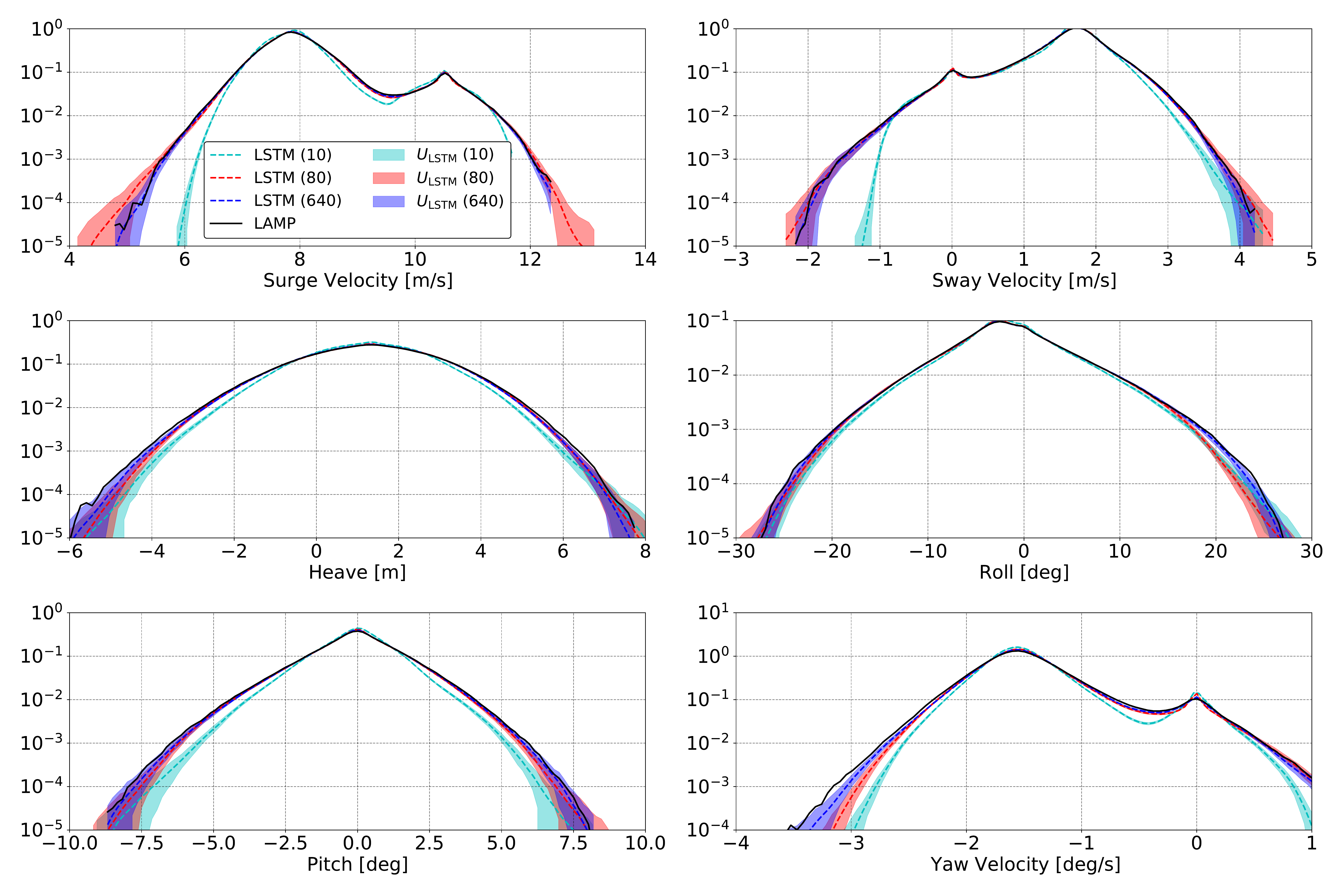}
		\caption{Comparison of the PDF tails for each DoF with models trained with 27 wave probes in the turning circle case study.}
		\label{fig:compareDistlog_turningCircle}
	\end{figure}

    The comparison between LAMP and the LSTM models for the PDF of each DoF are shown in Fig.~\ref{fig:compareDist_turningCircle} and in logarithmic scale in Fig.~\ref{fig:compareDistlog_turningCircle} for turning circle models built with 27~waves probes and training data quantities of 10, 80, and 640~runs.  Similar to the course-keeping case study, Fig.~\ref{fig:compareDist_turningCircle} shows that the PDF constructed from LSTM predictions closely match for the models trained with 80 and 640~runs. However, the model trained with 10~runs performs poorer than that of its course-keeping counterpart. The differences between the LSTM models in enhanced in the logarithmic scale shown in Fig.~\ref{fig:compareDistlog_turningCircle}, where increasing the amount of training data clearly results in a model converging towards the PDF predicted by LAMP. 
	
\begin{figure}[H]
		\centering
		\includegraphics[width=0.9\textwidth]{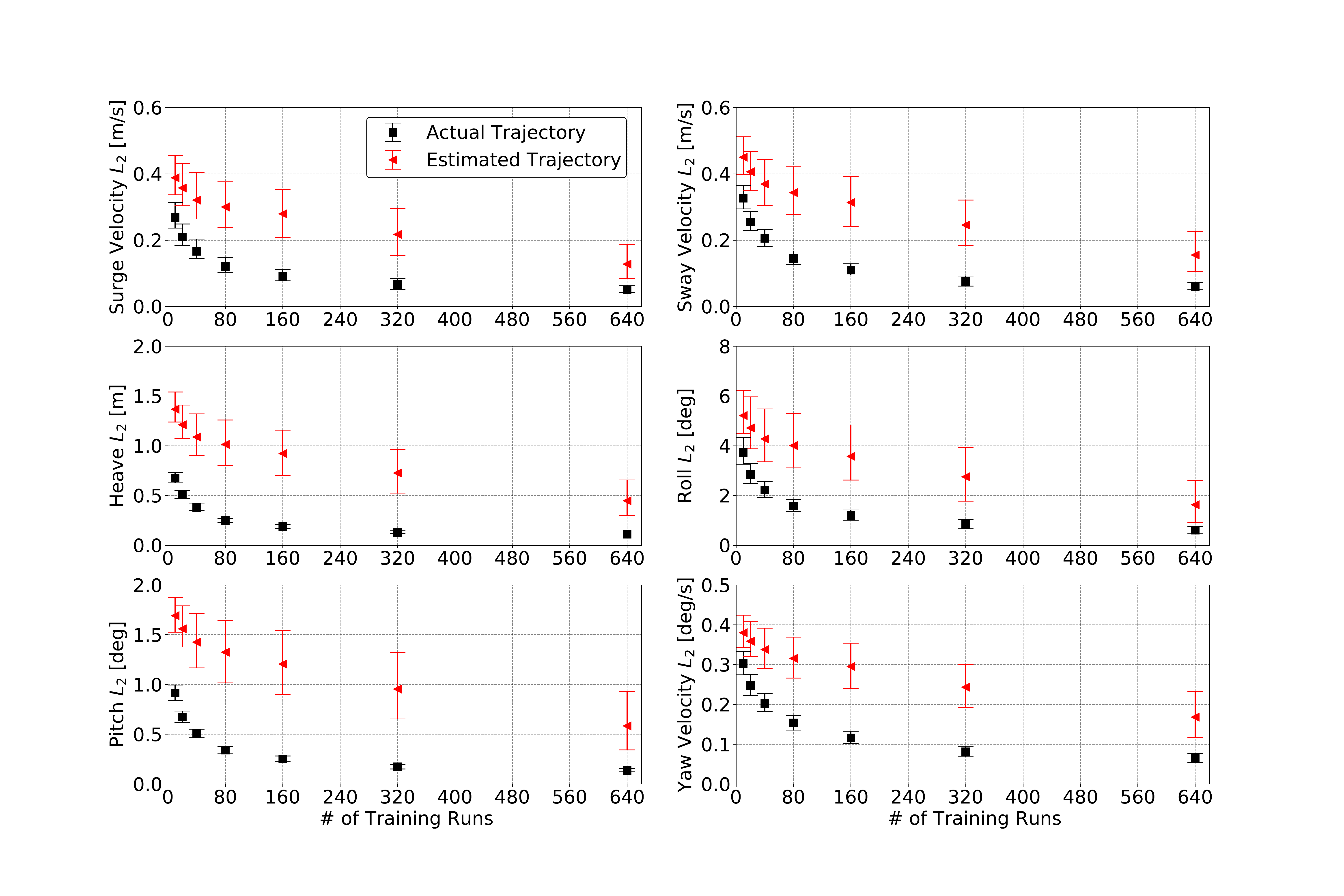}
		\caption{Comparison of $L_2$ error for each DoF with models trained with the actual and estimated encounter frames in the turning circle case study.}
		\label{fig:L2Compare_knownTrajectory_turningCircle}
	\end{figure}

\begin{figure}[H]
		\centering
		\includegraphics[width=0.9\textwidth]{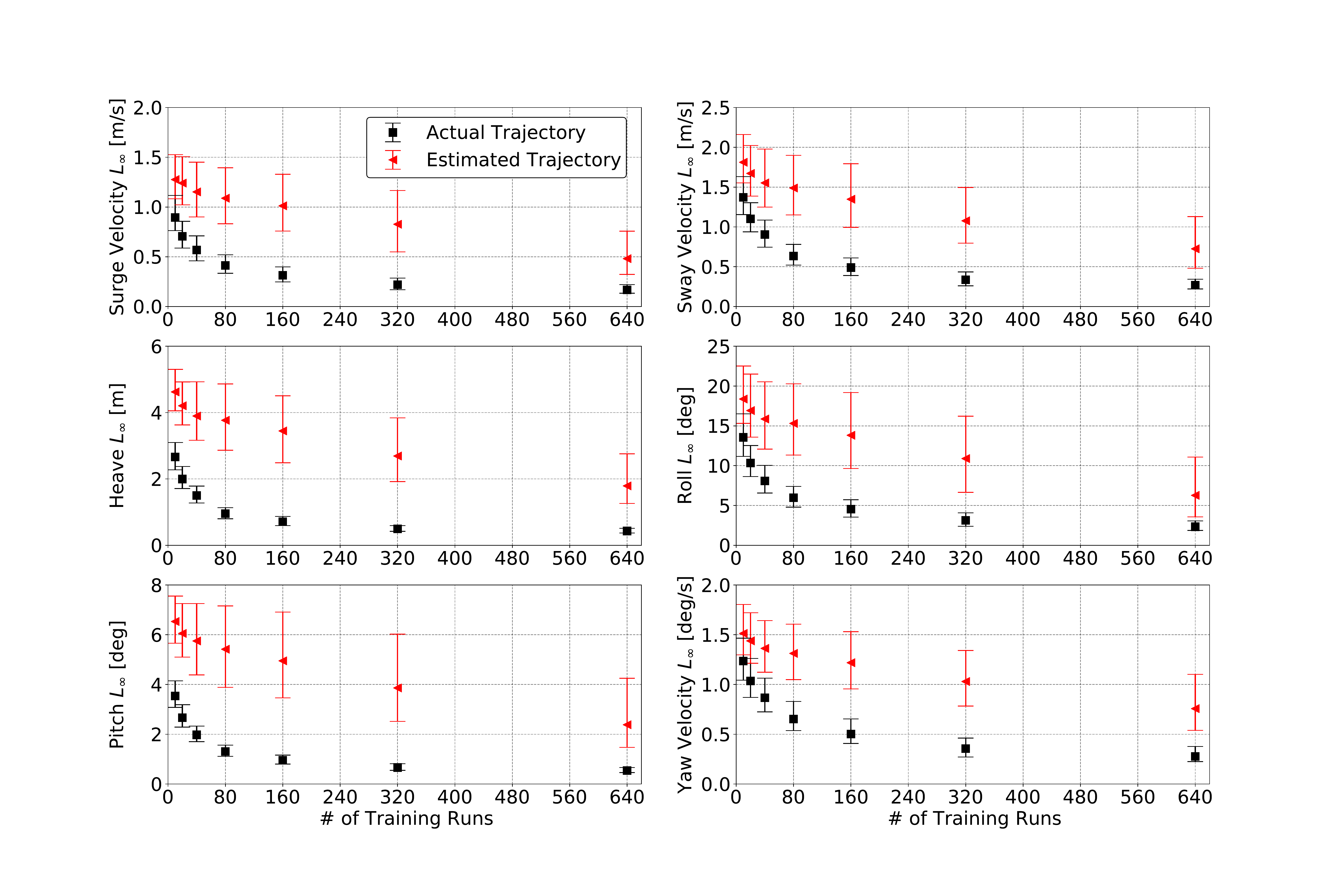}
		\caption{Comparison of $L_\infty$ error for each DoF with models trained with the actual and estimated encounter frames in the turning circle case study.}
		\label{fig:LinfCompare_knownTrajectory_turningCircle}
	\end{figure}

    In all the previous evaluations demonstrated for the turning circle LSTM models, the differences between LAMP and the LSTM predictions is shown again to be larger for turning circles than it was for course-keeping. Though, the higher frequency response could be the culprit, the encounter frame is also much more complicated for the turning circle as highlighted in Fig.~\ref{fig:estimateframeCourseKeeping} and \ref{fig:estimateframeTurning}. Fig.~\ref{fig:L2Compare_knownTrajectory_turningCircle} and \ref{fig:LinfCompare_knownTrajectory_turningCircle} show the comparisons of different LSTM models built with 27~wave probes and 640~training runs for turning circles, where the only difference is the encounter frame. Again, the difference are much more dramatic for the turning circle, where knowing the actual frame can greatly improve the predictions made by the LSTM model. This further indicates that a better prediction of the encounter frame will strongly benefit the model accuracy and greatly enhance the developed framework.
    	
	\section*{Conclusion}
	   
	A methodology is developed to represent the 6-DoF response of a free-running ship in waves with an LSTM neural network. An estimate of a mean instantaneous encounter frame is made from the training dataset and wave probes that travel with this estimated frame serve as input into the neural network, while the output is the 6-DoF response of the vessel. A Monte Carlo dropout technique is employed during training and model prediction to develop not only a more generalized model but also provide estimates of the model uncertainty during prediction. A comprehensive case study is performed with LAMP simulations of a free-running DTMB 5415 hullform operating at 20 knots in Sea State 7 stern-quartering irregular seas for both course-keeping and turning circles. Numerous neural network models are constructed and trained with various quantities of training data and wave probes as input. A clear convergence of models towards lower error is shown as the number of probes and training runs are increased. The LSTM neural network models are evaluated by calculating the $L_2$ and $L_\infty$ error for each model, and then comparing the validation run time histories with the largest and least amount of error for each DoF for the best performing neural network model. Additionally, the PDF of each DoF is compared between LAMP and the neural network predictions for models built with 27 wave probes and training data quantities of 10, 80, and 640 runs. The neural network predictions not only represent the overall PDF well but also the tail of the PDF is also accurately represented by the neural network models. 
	 
	 The present work demonstrates for the first time, that LSTM neural networks can be trained to represent the 6-DoF response of a free-running vessel in waves accurately. A rigorous and comprehensive case study has shown the methodology's effectiveness for accurately representing the non-trivial motions of the DTMB 5415 hull form operating at 20 knots in Sea State 7 stern-quartering seas. The presented case study is for a potential flow simulation tool, but the developed modeling approaches are applicable to experimental and full-scale data, as well as higher fidelity CFD simulations. In practice, a neural network could be pre-trained with high-fidelity CFD and deployed into the real-world where actual motion data could update the model based on different wave environments and thus improving the forecasting capabilities of the model. The work has implications in not only development of accurate surrogate models of ship motions but also in the real-time forecasting of vessel motions onboard both manned and unmanned vessels.

	\section*{Acknowledgments}
	
	This work is supported by the Department of Defense (DoD) Science, Mathematics, and Research for Transformation (SMART) scholarship, the Naval Surface Warfare Center Carderock Division (NSWCCD) Extended Term Training (ETT), and the NSWCCD Naval Innovative Science and Engineering (NISE) programs. The authors would also like to acknowledge and thank the Office of Naval Research for the support of this work under contracts N00014-20-1-2096 by the program manager Woei-Min Lin.

	\bibliographystyle{abbrvnat}
	\bibliography{References}  
	
\end{document}